\documentclass{article}

\usepackage[preprint,nonatbib]{neurips_2023}

\usepackage[utf8]{inputenc} 
\usepackage[T1]{fontenc}    
\usepackage{hyperref}       
\usepackage{url}            
\usepackage{booktabs}       
\usepackage{amsfonts}       
\usepackage{nicefrac}       
\usepackage{microtype}      
\usepackage{xcolor}         

\usepackage{graphicx}
\usepackage{amssymb}
\usepackage{booktabs}
\usepackage{amsmath,amssymb,amsfonts}
\usepackage{amsbsy}
\usepackage{graphics}
\usepackage{algorithm}
\usepackage{algpseudocode}
\usepackage{caption}
\usepackage{subcaption}
\usepackage{multirow}
\usepackage{ctable}
\usepackage{siunitx}
\usepackage[capitalize]{cleveref}
\crefname{section}{Sec.}{Secs.}
\Crefname{section}{Section}{Sections}
\Crefname{table}{Table}{Tables}
\crefname{table}{Tab.}{Tabs.}

\newtheorem{claims}{{Claim}}
\newtheorem{claim}[claims]{{Claim}}
\DeclareMathOperator\argmax{argmax}
\DeclareMathOperator\argmin{argmin}

\DeclareMathOperator\smoothmax{smoothmax}
\DeclareMathOperator\erf{erf}

\title{Regularizing Differentiable Architecture Search with Smooth Activation}

\author{Yanlin Zhou, Mostafa El-Khamy, Kee-Bong Song\\
Samsung Semiconductor, Inc.\\
9868 Scranton Road, San Diego, CA, USA\\
{\tt\small \{yanlin.z, mostafa.e, keebong.s\}@samsung.com}
}

\begin{document}

\maketitle

\begin{abstract}
Differentiable Architecture Search (DARTS) is an efficient Neural Architecture Search (NAS) method but suffers from robustness, generalization, and discrepancy issues. 
Many efforts have been made towards the performance collapse issue caused by skip dominance with various regularization techniques towards operation weights, path weights, noise injection, and super-network redesign.
It had become questionable at a certain point if there could exist a better and more elegant way to retract the search to its intended goal -- NAS is a selection problem.
In this paper, we undertake a simple but effective approach, named \textit{Smooth Activation DARTS} (SA-DARTS), to overcome skip dominance and discretization discrepancy challenges.
By leveraging a smooth activation function on architecture weights as an auxiliary loss, our SA-DARTS mitigates the unfair advantage of weight-free operations, converging to fanned-out architecture weight values, and can recover the search process from skip-dominance initialization.
Through theoretical and empirical analysis, we demonstrate that the SA-DARTS can yield new state-of-the-art (SOTA) results on NAS-Bench-201, classification, and super-resolution.
Further, we show that SA-DARTS can help improve the performance of SOTA models with fewer parameters, such as Information Multi-distillation Network on the super-resolution task.  
\end{abstract}

\section{Introduction}

Differentiable Architecture Search (DARTS) \cite{liu2018darts} has been dominating the field of neural architecture search (NAS) for its efficient gradient-based paradigm achieved by the continuous relaxation of candidates and the one-step approximation of bi-level optimization.

DARTS suffers from two issues: skip dominance and discrepancy of discretization \cite{chu2020darts, liang2019darts+, wang2021rethinking, zela2019understanding}.
Thus, DARTS suffers from performance collapse issues caused by aggregation of skip-connections, and the searched architectures prefer to accumulate parameter-free operations \cite{liang2019darts+, zela2019understanding}.
Early stopping \cite{zela2019understanding}, path-regularization \cite{zhou2020theory}, dropout for skip-connections \cite{chen2019progressive}, auxiliary skip-connection \cite{chu2020darts}, increasing weight decay, and limiting skip-connection \cite{chu2021fairnas} allowance are many practical tricks to help stabilize the search.
These methods focus on manually designing more rules for NAS, which contradicts the ultimate goal of automated search. 

Other work focus on better optimization of supernet, including early stopping \cite{zela2019understanding}, hand-crafted constraints \cite{chu2021fairnas}, or smoothing the validation loss landscape \cite{chen2020stabilizing}.
Hessian eigenvalue is also used as an indicator to stop the searching before the performance drops \cite{chen2020stabilizing, chen2020drnas, zela2019understanding}.

\begin{figure*}[tpbh]
    \centering
    \includegraphics[width=\textwidth]{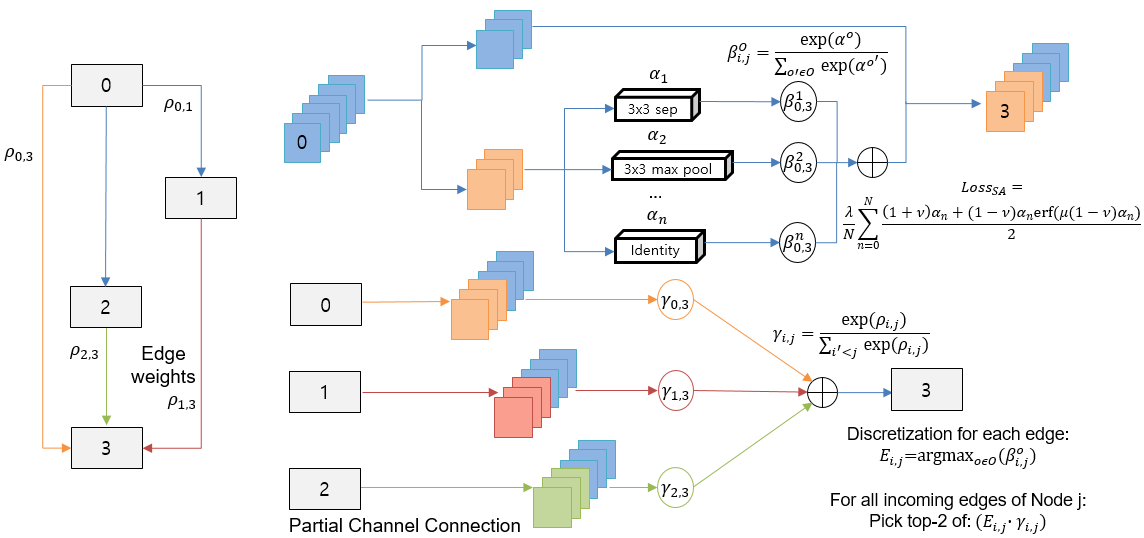}
    \caption{Overflow of SAC-DARTS. In addition to the cross-entropy loss, the auxiliary loss $L_{SA}$ is added as a regularization term. The DARTS-based cell is shown on the left. Within each cell, only a portion of the channels is selected for mixed operation, while unselected channels are directly passed to the next node. Edge weights are introduced to reduce stochastic incurred by channel sampling. At the final child network discretization step, top-2 incoming edges with the highest $\alpha$ values are kept.}
    \label{fig:mr-pcdarts}
\end{figure*}

Another comparably less studied open problem is the discrepancy issue due to the gap between the performance of the over-parameterized supernet and its final derived child network \cite{chu2020fair}.
During DARTS's supernet search phase, all operations are used between feature maps in a weight-sum manner.
When deriving the final network, all but one operations are pruned between connected feature maps, keeping the operation with the most significant contribution in a supernet.
In most cases, the magnitude of operation contributing weights does not genuinely reflect the rank of operations \cite{wang2021rethinking}. 
Thus, there is a performance gap between the supernet and its derived child network.
This pruning behavior introduces a discrepancy from continuous supernet representation to discrete child one-hot encoding.
Further, input perturbation is effective \cite{chen2020stabilizing}.
However, the ultimate goal of DARTS is to find the best architecture rather than explicitly dealing with skip-connection. 
Regardless of how successful those regularization tricks (\textit{i.e.,} weight-decay \cite{zela2019understanding}, early stopping \cite{liang2019darts+}) excel in minimization of supernet losses, provide proper operation ranking is a very different task. 
Because, \textit{NAS is a selection problem.}

In this paper, we present smooth activation DARTS (SA-DARTS) with a rather unorthodox auxiliary loss.
We regularize the update of architecture weights $\boldsymbol {\alpha}$ of all candidate operations by exerting a smooth activation function on $\boldsymbol {\alpha}$ as a loss term.
The design ethos of SA regularization easily extends to other DARTS-like algorithms.
Our SA-DARTS not only overcomes skip-dominance issues but also solves the discrepancy issue where other SOTAs fail.
SA-DARTS allows operators to take dispersed architecture weights without the need for engineering tricks like a warm-up.
Additionally, SA-DARTS is more robust against local optimal during the search phase and can quickly recover from undesired conditions, \textit{i.e.,}, skip carrying a higher weight than other operations.
We further speed up our method with partial-channel \cite{xu2021partially} and name it SAC-DARTS.
Through extensive experimental results over different vision tasks, \textit{e.g.,}, classification and super-resolution, both SA-DARTS and SAC-DARTS achieve SOTA performances. 
Further experiments on NASBench201 show that SA-DARTS reduces the discrepancy between supernet architecture encoding and the derived one.

\begin{figure}[tpbh]
     \centering
     \begin{subfigure}[b]{0.65\textwidth}
         \centering
         \includegraphics[width=\textwidth]{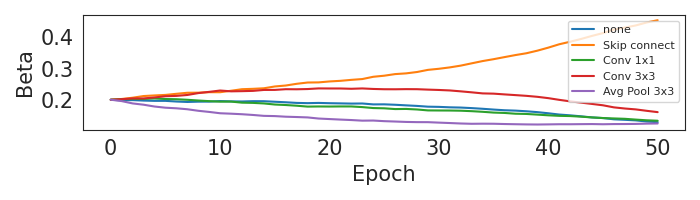}
         \caption{DARTS with L2 regularization}
     \end{subfigure}    
     \begin{subfigure}[b]{0.65\textwidth}
         \centering
         \includegraphics[width=\textwidth]{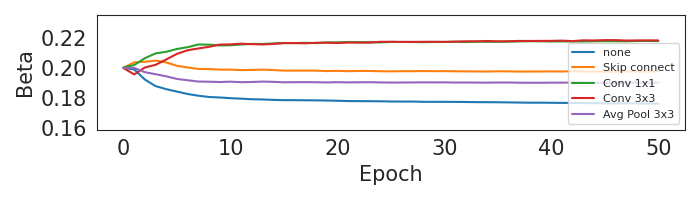}
         \caption{Beta-DARTS}
     \end{subfigure}   
     \begin{subfigure}[b]{0.65\textwidth}
         \centering
         \includegraphics[width=\textwidth]{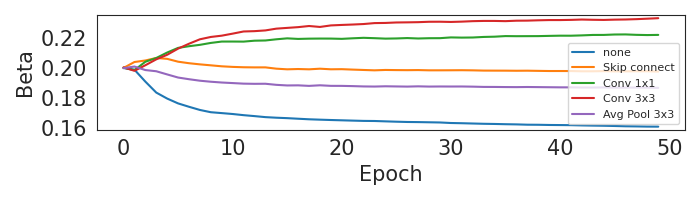}
         \caption{Our SA-DARTS}
     \end{subfigure}
        \caption{Comparison between DARTS with L2 regularization ($\lambda_2 \sum^{|O|}_{k=1} \alpha_k^2$), Beta regularization ($\lambda_\beta \log(\sum^{|O|}_{k=1} e^{\alpha_k})$), and our Smooth Activation regularization of their $\beta$ (softmax of architecture weights $\alpha$) change vs. epochs on a same edge. While the original DARTS with L2 in (a) suffers from the curse of skip dominance, and the top-2 softmax values of Beta-DARTS in (b) are too close to identify a suitable candidate operation, our SA-DARTS (c) solves the skip dominance issue and achieves dispersed $\beta$ values and alleviates the discrepancy of discretization issue. The search is done on NAS-Bench-201.}
        \label{fig:darts_beta_mr_beta}
\end{figure}


\section{Background and Problem Formulation}\label{sec:regularization}

\subsection{Preliminary of DARTS}

DARTS searches for repeatedly stacked cells to construct a convolutional neural network (CNN).
Each computation cell is a directed acyclic graph with 7 nodes: 2 input nodes from the immediate previous cells, 4 intermediate nodes, and an output node.
Each node $X_i$ is a feature map, and each directed edge $(i,j)$ between nodes contains $N$ operations to transform $X_i$ to $X_j$.
Four intermediate nodes are densely connected with predecessors as $x_j = \sum_{i<j} \overline{o}^{(i,j)}(x_i)$.
The candidate operations include: original/separable/dilated convolutions with different filter size, pooling, skip, none, \textit{etc.}

To make the search space continuous, DARTS relaxes the categorical choice of a particular operation to a softmax over all possible operations.
DARTS defines architecture variables $\alpha$ as indicators for each operator's contribution.
After the continuous relaxation, the goal is to solve the bi-level optimization problem by jointly learning $\alpha$ and its optimal CNN weight $w$ on validation and training dataset separately, $\min_\alpha L_{val}(w^*(\alpha), \alpha) \\
    \text{s.t. } w^*(\alpha)= \argmin_w L_{train}(w, \alpha) $.
The weight is measured
\begin{equation}\label{eqn:beta}
    \beta = \frac{\exp(\alpha_o^{i,j})}{\sum_{o'\in O}\exp(\alpha_{o'}^{i,j})}
\end{equation}
Then the mixed operation is defined as:

$\overline{o}^{(i,j)}(x) = \sum_{o \in O}\frac{\exp(\alpha_o^{i,j})}{\sum_{o'\in O}\exp(\alpha_{o'}^{i,j})} o(x) $,
where $O$ is operation set, and $o(x_i)$ is a selected operation to be applied on $x_i$. 
The operation mixing weights for a pair of nodes $(i,j)$ are parameterized by a vector $\alpha^{(i;j)}$ of dimension $|O|$.
Cells of the same type (\textit{e.g.,}, normal or reduction) share the same $\alpha$.
As a result, each edge between nodes is a weighted sum $\overline{o}$ of all 8 operations.
The task of architecture search then reduces to learning a set of continuous $\alpha$ variables -- the encoding of architectures -- to be optimized.
After done searching, the final network can be obtained by replacing mixed operation $o^{(i;j)}$ on each edge with the most likely candidate.
Each node first selects the best operation within each edge, then picks the top-2 among all incoming input edges, according to $\alpha$.
Only 12 out of 14 edges are kept for the final network.
\begin{equation}\label{eqn:dartsargmax}
    o^{(i;j)} = \argmax_{o \in O} \alpha_o^{(i;j)}
\end{equation}



\begin{figure*}[tpbh]
     \centering
     \begin{subfigure}[b]{0.49\textwidth}
         \centering
         \includegraphics[width=\textwidth]{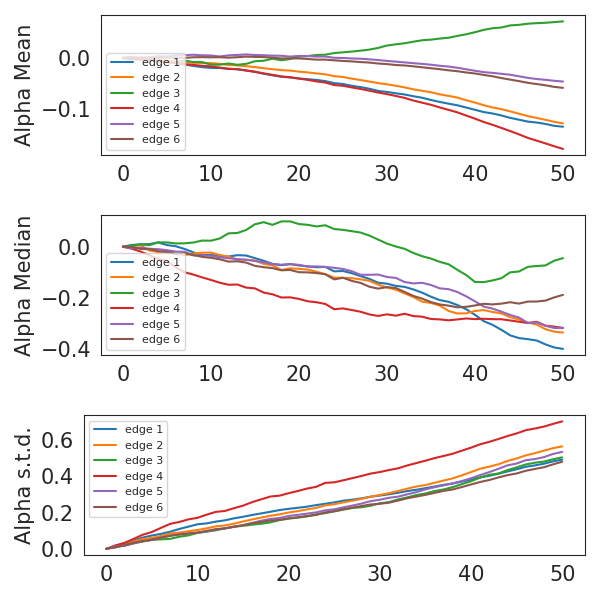}
         \caption{L2 Regularization.}
     \end{subfigure}
     \begin{subfigure}[b]{0.49\textwidth}
         \centering
         \includegraphics[width=\textwidth]{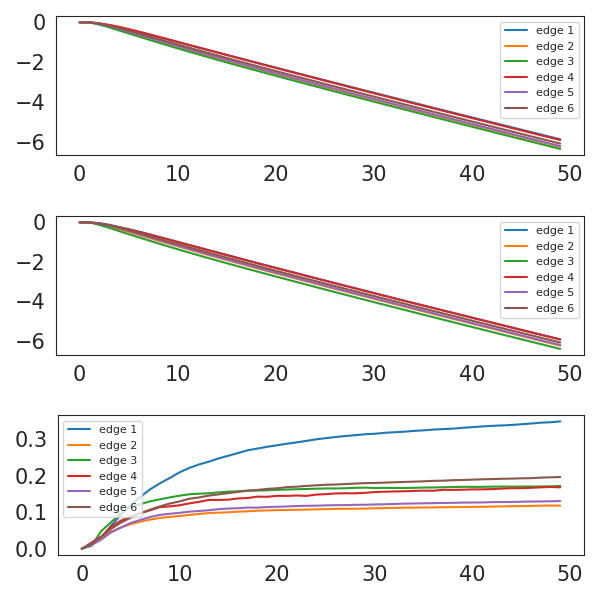}
         \caption{Our Smooth Activation Regularization.}
     \end{subfigure}
        \caption{Comparison between different regularization of DARTS: $L_2 = \lambda_2 \sum^{|O|}_{k=1} \alpha_k^2$ and our $L_{SA}=\dfrac{\lambda}{N} \sum_{i=1}^{N}  \dfrac{(1+\nu)\alpha_i+(1-\nu)\alpha_i \erf(\mu(1-\nu)\alpha_i)}{2}$ of their $\alpha$ mean  vs. epoch, $\alpha$ median  vs. epoch, and $\alpha$ standard deviations vs epochs. To solve skip dominance issue, our SA-DARTS drives $\alpha$ values to large negative numbers so their change in softmax is relatively smaller when compared to positive $\alpha$ values. The search is done with NAS-Bench-201.}
        \label{fig:mean_median_std}
\end{figure*}

\subsection{Eliminating the Skip Dominance}\label{sec:peformancecollapse}

The original DARTS method suffers from the skip dominance issue reported in \cite{zela2019understanding}, and the final network usually converges to all skip networks.
Without loss of generality, following \cite{wang2021rethinking}, let us consider a simplified search space for one cell that consists of three operations: 3x3 convolution, skip, and average pooling.
The softmax of architecture parameter $\alpha_{conv}, \alpha_{skip}, \alpha_{avg}$ follow that $\beta_{conv} + \beta_{skip} + \beta_{avg} = 1$.
Let $x$ be the input feature map, $o_c(x)$ be the output of convolution operation, $o_a(x)$ be the output of average pooling, and $m^*$ be the optimal output feature map.
The current estimation of the optimal output feature map becomes the mixed operation: 
$ \overline{m} (x) = \beta_{conv}o_c(x) + \beta_{skip} x + \beta_{avg} o_a(x) $.
The objective can be formulated as: 
$   \text{min}_{\beta_{conv}, \beta_{skip}, \beta_{avg}} \text{Var}(\overline{m} (x) - m^*)$.

Following the 2 operator case in \cite{wang2021rethinking}, we can group the convolutional and average pooling, and proof by recursion:

\begin{equation}\label{eqn:alphaconv}
     \alpha_{conv}^* \propto \text{Var}(x-m^*) + \text{Var}(o_a(x)-m^*)
\end{equation}
\begin{equation}\label{eqn:alphaskip}
     \alpha_{skip}^* \propto \text{Var}(o_c(x)-m^*) + \text{Var}(o_a(x)-m^*)
\end{equation}
\begin{equation}\label{eqn:alphaavg}
     \alpha_{avg}^* \propto \text{Var}(o_c(x)-m^*) + \text{Var}(x-m^*)
\end{equation}

Equation~\ref{eqn:alphaconv} - \ref{eqn:alphaavg} reveal the skip-connection dominance issue because $\alpha_{skip}$'s magnitude is accumulated by two greater variance terms.
While $x$ comes from the mixed output of the previous edge, it will naturally be closer to $m^*$, $o_a(x)$ is an average pooling, which will be slightly off but very consistent, and $o_c(x)$ is the output of a single convolution operation, which is inconsistent and tend to deviate more from $m^*$ during training, becoming the furthest.
From an empirical perspective, in the early stage, the search algorithm often prefers weight-free operations, especially \textit{skip-connection}, because no training is involved, and the outputs are more consistent.
In contrast, the weight-equipped operations would propagate inconsistent information across iterations before their weights are well-optimized.
Consequently, the \textit{skip-connection} often accumulates dominant weights at the beginning, which makes it difficult for the other operations to be selected later.
The update of architecture weights $\boldsymbol {\alpha}$ should be close to each other at the early searching stage.
There is a pitfall of using weight decay \cite{zela2019understanding} and L2 \cite{liu2018darts} to regularize $\alpha$.
The Adam optimizer with L2 is commonly used for DARTS, but the gradients of L2 regularization are normalized by their summed magnitudes. 
Thus the penalty for each element is relatively even.
Therefore, \cite{ye2022b} adopt Log Sum Exponent (LSE), $ L_\beta = \lambda  \cdot \log(\sum^{|O|}_{k=1} e^{\alpha_k}) = \lambda \cdot \smoothmax({\alpha_k})$, as an additional loss term to mitigate above issue and constrain the value of $\beta$ from changing too much \cite{ye2022b}.

\subsection{Eliminating the Discrepancy of Discretization}\label{sec:discrepancy}

Several papers challenge \cite{chu2020fair, zhou2020theory} the assumption that the $\alpha$ truly reflects the strength of the underlying operations. 
Specifically, it happens after an over-parameterized supernet is trained with a set of alpha values for each edge.
The performance of the supernet is very different from the child network derived using $\argmax$.
It is concluded that $\alpha$ does not necessarily result in higher validation accuracy after discretization \cite{wang2021rethinking}.

In addition, FairDARTS \cite{chu2020fair} stated the discrepancy of discretization is due to $\beta$ being too close to each other when making the final one-hot choice.
Recall that each edge in DARTS uses all eight operations by weighted sum during the search phase and discretizes the child network using Equation~\ref{eqn:dartsargmax}.
In most cases, $\alpha$ values of operations are very close.
For instance, FairDARTS shows one of their edges contains the following $\beta$ values $[0.176, 0.174, 0.170,...]$. 
It is a good example to show that the $\beta$ range is too narrow to identify a good operation.
However, in Beta-DARTS \cite{ye2022b}, $\beta$ values of different operations are very close to each other.
We show in Section \ref{sec:sr} that Beta-DARTS alternates among a few candidates in super-resolution tasks. 
The continuous encoding of DARTS architecture weights $\boldsymbol {\alpha}$ should reflect actual one-hot discretization with wider $\boldsymbol {\alpha}$ differences.



\section{Methodology}

Consider the simplified search space in Section~\ref{sec:peformancecollapse}, it can be easily inferred that the $\alpha_{skip}$, $\alpha_{conv}$, and $\alpha_{avg}$ follow:

\begin{multline}\label{eqn:alpha_conv}
    e^{\alpha_{conv}} + e^{\alpha_{skip}} + e^{\alpha_{avg}} = 2*\text{Var}(x-\overline{m} (x)) + \\ 2*\text{Var}(o_a(x)-\overline{m} (x)) + 2*\text{Var}(o_c(x)-\overline{m} (x)) - \\ 3 * \text{Cov}(x-\overline{m} (x), o_c(x)-\overline{m} (x), o_a(x)-\overline{m} (x)) + C
\end{multline}
Therefore, to reduce the overall variances, we can choose a smaller value for $\alpha$, so $\exp(\alpha)$ is small.



\begin{algorithm}[t]
    \caption{SAC-DARTS}
    \label{alg:pcb}
    \begin{algorithmic}[1]
	\State Initialize architecture parameter $\alpha$; edge weights $\gamma$; Network weights $w$; Number of search epochs $E$;
	\State Construct a supernet and initialize architecture parameters $\alpha$, edge weights $\gamma$, and supernet weights $w$;
	\State Create a mixed operation $\Bar{o}^{i,j}$ parameterized by $\alpha^{(i,j)}$ for each edge $(i,j)$
	\For{each epoch $e \in [1, E]$} 
		\State 
		\State Update architecture parameters $\alpha$ by descending first-order $\triangledown_{\alpha} L_{val}(w,\alpha) + \lambda_e L_{SA}$
		\State Update network weights $w$ by descending $\triangledown_{w}L_{train}$ 
	\EndFor
	\State Finally, DERIVE() the final architecture.
	\State
	\Function{Derive}{$\alpha$, $\gamma$, $OPS$}
		\State For each edge between nodes, calculate the selection value $v = \alpha * \gamma$, and keep the operation with the $\argmax_{op \in OPS}(\alpha_{op} * \gamma_{op})$, and delete the rest operations.
		\State For each node in cell, keep 2 incoming edges: $ o^{(i,j)} = \argmax_{o \in O}(\alpha_o^{(i,j)})$
		\State \Return structure list
	\EndFunction
	\State
	\Function{Mixed Operation}{$x$, $OPS$, $\beta$, $k$}
		\State Shuffle channels then divide input $x$ to $x_1$ and $x_2$ according to partial channel ratio $k$.
		\State Apply all operations $OPS$ on $x_1$ with weights $\beta$:  $x_1' = \sum_{op \in OPS}(\beta_{op} * op(x_1))$
		\State \Return Concate($x_1'$, $x_2$)
	\EndFunction
	\end{algorithmic}
\end{algorithm}


To both eliminate the unfair advantage of skip-connection and introduce a bigger dispersion among the softmax value of architectural weights, 
we approach this problem from a different perspective and encapsulate the $\alpha$ terms from Equation~\ref{eqn:alpha_conv} inside an activation function.
Since $\alpha$ can take both positive or negative values, Leaky RELU would be a great initial choice.
However, we need a smooth approximation \cite{biswas2021smu} of general Maxout family, so we can make this term differentiable and extend to a wider selection of Maxout Unit for various vision tasks with different hyperparameters.

We introduce our smooth activation regularization term in addition to DARTS's cross entropy loss :
\begin{equation}\label{eqn:smu}
\resizebox{0.5\textwidth}{!}{$
    L_{SA} = \dfrac{\lambda_e}{N_o N_e} \sum_{i=1}^{N_e} \sum_{j=1}^{N_o}  \dfrac{(1+\nu)\alpha_i^j+(1-\nu)\alpha_i^j \erf(\mu(1-\nu)\alpha_i^j)}{2}$
}
\end{equation}
where $N_o, N_e$ are number of candidate operations and edges, 
$\lambda_e$ is a linearly increasing coefficient to control the loss strength,
$\erf()$ denotes the Gaussian error function,
$\nu$ and $\mu$ are hyperparameters that determine the maxout unit type.
For instance, as $\mu \rightarrow \infty$, Equation~\ref{eqn:smu} approximates ReLU or Leaky ReLU depending on the value of $\nu$.
During our experiments, there exits multiple pairs of $\nu$ and $\mu$ that can achieve SOTA on different vision tasks.
If we set $\nu = 1$ or $\mu = 0$, Equation~\ref{eqn:smu} reduces to regularizing mean values.
\begin{equation}\label{eqn:mrloss}
    L_{SA} = \dfrac{\lambda_e}{N_o N_e} \sum_{i=1}^{N_e} \sum_{j=1}^{N_o} \alpha_i^j  
\end{equation}
We choose $\lambda_e = (\text{epoch}/5)$ for most experiments, as it generalizes the best.
If not mentioned otherwise, we refer to the special case in Equation~\ref{eqn:mrloss} as default.
Our SA-DARTS only introduce negligible computation cost by adding an additional loss term to the DARTS searching phase.

Furthermore, we propose to leverage the partial channel trick \cite{xu2021partially} on SA-DARTS to aim for an efficient search method while achieving the SOTA performance.
We name the new model smooth activation channel DARTS (\textbf{SAC-DARTS}), as shown in Fig.~\ref{fig:mr-pcdarts}.
We regularize the operation weights of DARTS, but not the edge weights by PC-DARTS.
With DARTS search space, we have 14 edges in each cell, each with 8 operations for both normal and reduction cells.

\subsection{Comparison of DARTS with L2, LogSumExp, and Smooth Activation Regularization}

\begin{table}[tpbh]
    \centering
    \begin{tabular}{|l|c|c|}
    \hline
        Method & Converge &  Difference \\
    \hline
        DARTS  \cite{liu2018darts} & No &  $\num{6E-2} \pm \num{3E-2}$  \\
        Beta-DARTS \cite{ye2022b} & Yes & $\num{1E-3} \pm \num{5E-4}$ \\
        SA-DARTS    &  Yes &    $ \num{1E-2} \pm \num{2E-3}$\\
    \hline
    \end{tabular}
    \caption{Comparison of DARTS with L2, Beta-DARTS, and SA-DARTS of their $\beta$ dispersion as an indicator of discrepancy of discretization. We measure the top-2 contributing operations' $\beta$ at $50^{th}$ searching epoch on NAS-Bench-201. The bigger, the better. The results are generated across 10 trials on the same edge with the closest difference between top-2 $\alpha$.}
    \label{tab:beta-dispersion}
\end{table}

DARTS clearly has skip dominance issue as shown in Fig.~\ref{fig:darts_beta_mr_beta} (a).
However, our SA-DARTS drives $\alpha$ to negative numbers as shown in Fig.~\ref{fig:mean_median_std}(b), and the leading privilege of $\beta_{skip}$ is constrained.
For the SA-DARTS case in Fig.~\ref{fig:darts_beta_mr_beta} (c), skip-connection (in orange) was unquestionably in a leading place before epoch 5, but was later successfully constrained in later search epochs.
Our method is equivalent to shifting exponential calculation in Equation~\ref{eqn:beta} to a more stable output range, so the accumulation of skip-connection only increases linearly.
Moreover, SA-DARTS also brings stable standard deviation values of $\alpha$, as shown in Fig.~\ref{fig:darts_beta_mr_beta} (c).
Thus, the $\beta$ probability updates are steadier according to Equation~\ref{eqn:alphaconv} - \ref{eqn:alphaavg}, granting a stable search process.
Our SA regularization is necessary, rather than relying on negative $\alpha$ initialization.
we also experiment that the original DARTS with \textit{negative initialization} \{-5, -2, -1, -0.5\} for all $\alpha$ still experiences skip dominance issue, and cannot achieve the similar effect of regularization.

\begin{claim}
SA-DARTS overcomes the skip dominance issue by pushing architecture parameter variables $\alpha$ to large negative values, thus resulting in gentle changes of $\beta$ and enabling better generalization.
\end{claim}

In addition to overcoming skip dominance issues, we demonstrate that SA-DARTS mitigates the discrepancy issue implicitly.
The special case of SA regularization in Equation~\ref{eqn:mrloss} loosely regularizes the mean of all $\alpha$ values.
SA allows the gradient magnitude of $\alpha$ to be adequately small to fluctuate, but large enough to discern the top contributing candidate.
Recall that Beta-DARTS suppresses any greater value using LSE and making all $\beta$ values close. On the other hand, our SA in Equation~\ref{eqn:mrloss} regularizes the mean value to accommodate larger and smaller $\beta$ values in a quasisymmetric manner.

SA-DARTS aims to increase the spread between $\beta$ values, without enforcing any particular ranking, to enhance the argmax selection.
Why dispersed $\beta$ values?
Selecting the best operator through the $\beta$ probability is analogous to \textbf{maximizing the classification margin} under cross-entropy loss-based probabilities.
Cross-entropy loss does not inherently optimize for margin. 
Incorporating additional loss terms that push for larger margin solutions can improve generalization and robustness.

To verify the consistency in combating the discrepancy issue, Table~\ref{tab:beta-dispersion} compares the top-2 $\beta$ values for original DARTS, Beta-DARTS, and SA-DARTS. 
We collect 10 results for each method concerning their worst edge in each trial.
It shows that SA-DARTS is better at identifying the best candidate at a more considerable margin.
Further, we compare the accuracy of loss landscape of SA-DARTS vs. original DARTS and show that SA-DARTS can land on a smoother/flatter validation loss/accuracy landscape compared to original DARTS.

\begin{claim}
SA-DARTS overcomes the discrepancy of discretization issue by regularizing the mean of $\alpha$ to accommodate larger and smaller $\beta$ values in a quasisymmetric manner. The resulting continuous encoding of SA-DARTS is closer to the exclusive discrete one-hot encoding.
\end{claim}


\begin{figure}[tpbh]
     \centering
     \begin{subfigure}[b]{0.49\textwidth}
         \centering
         \includegraphics[width=\textwidth]{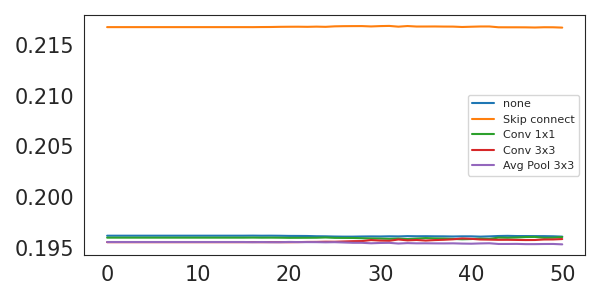}
         \caption{DARTS.}
         \label{fig:unfairmr}
     \end{subfigure}
     \hfill
     \begin{subfigure}[b]{0.49\textwidth}
         \centering
         \includegraphics[width=\textwidth]{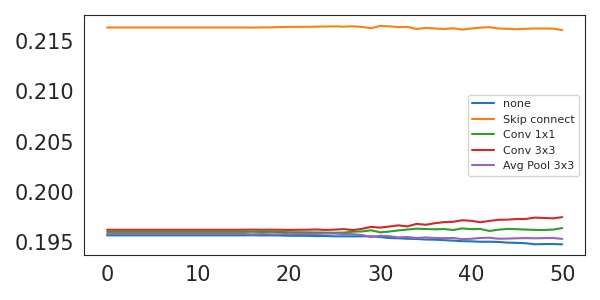}
         \caption{Beta-DARTS.}
         \label{fig:unfairdarts}
     \end{subfigure}
     \hfill
     \begin{subfigure}[b]{0.6\textwidth}
         \centering
         \includegraphics[width=\textwidth]{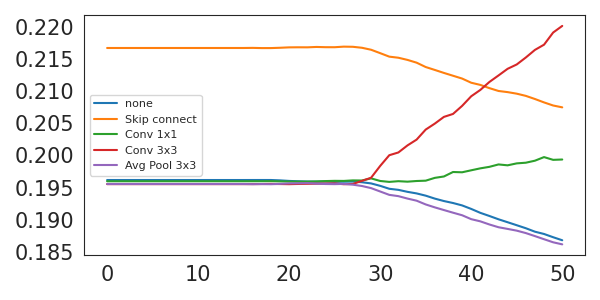}
         \caption{SAC-DARTS.}
         \label{fig:unfairbeta}
     \end{subfigure}
        \caption{$\beta$ value changes of the same edge with L2, Beta, and SA on DARTS at an unfair local optimal favoring skip-connection. We assign skip-connection with a higher probability. The first 15 epochs are for warm-up only. Our SAC-DARTS can recover from the unfair disadvantage and jump out of the local optimal. The search is done on NAS-Bench-201.}
        \label{fig:unfair}
\end{figure}

\subsection{Recovering from Unfair Disadvantage}

We show in Fig.~\ref{fig:unfair} that SA-DARTS helps the search return from an unfair state.
To further test the robustness of SA, L2, and Beta regularization, we consider a challenging search case where skip-connection already carries a higher weight than all other operations.
We choose to verify results using NAS-Bench-201 quickly.
We initialize the $\alpha$ to be [0, 0.1, 0, 0, 0], where $0.1$ is added to skip-connection.
The adjusted initial $\beta$ becomes [0.196, 0.216, 0.196, 0.196, 0.196].
Recall that the skip-connection is naturally preferred during the search phase, as proved in Section~\ref{sec:peformancecollapse}, now the condition is worse with uneven weights.
We set the first 15 epochs for warm-up only, a trick to stabilize DARTS, so we know the baselines are at their full potential.
With the original DARTS, or even recent SOTA Beta-DARTS, the skip dominance cannot be mitigated, as shown in Fig.~\ref{fig:unfair}.
However, our SAC-DARTS can recover from this extremely unfair competition between the \textit{skip-connection} and \textit{conv 3x3}.
Eventually, \textit{conv 3x3} (in red, also ground truth) overcomes the skip dominance issue,  receives a greater update signal, and becomes the distinct top candidate.

\begin{figure*}[thbp]
     \centering
     \begin{subfigure}[b]{0.49\textwidth}
         \centering
         \includegraphics[width=\textwidth]{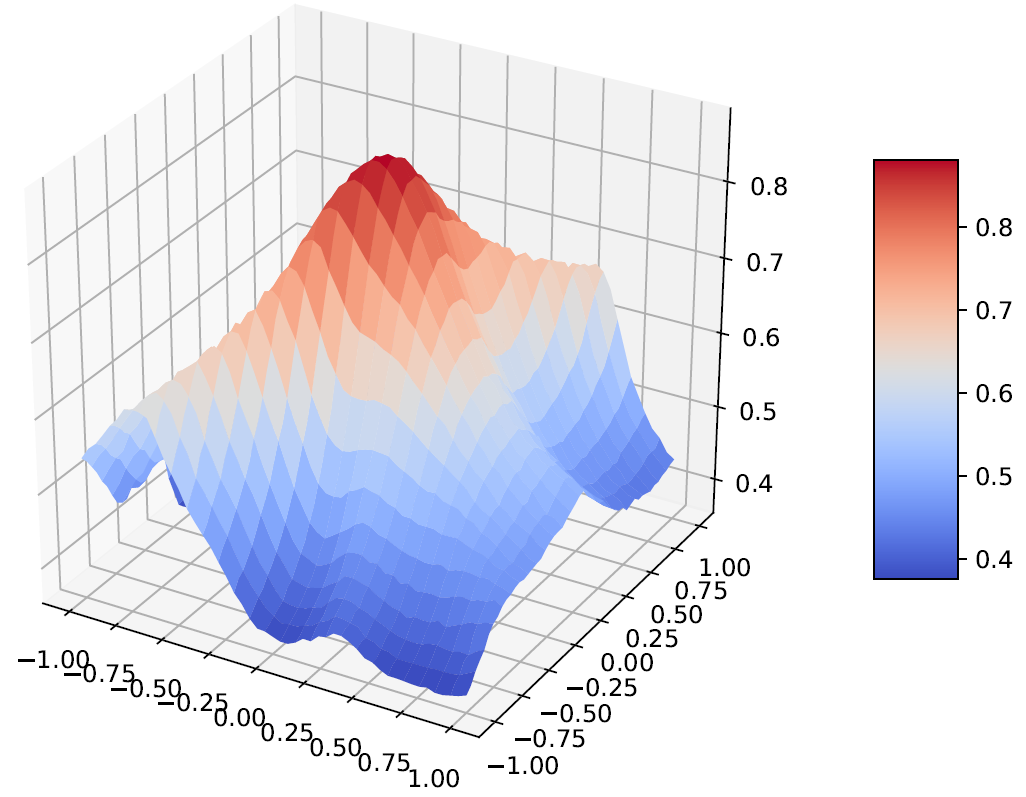}
         \caption{DARTS validation accuracy landscape.}
         \label{fig:landscape-sharpacc}
     \end{subfigure}
     \begin{subfigure}[b]{0.49\textwidth}
         \centering
         \includegraphics[width=\textwidth]{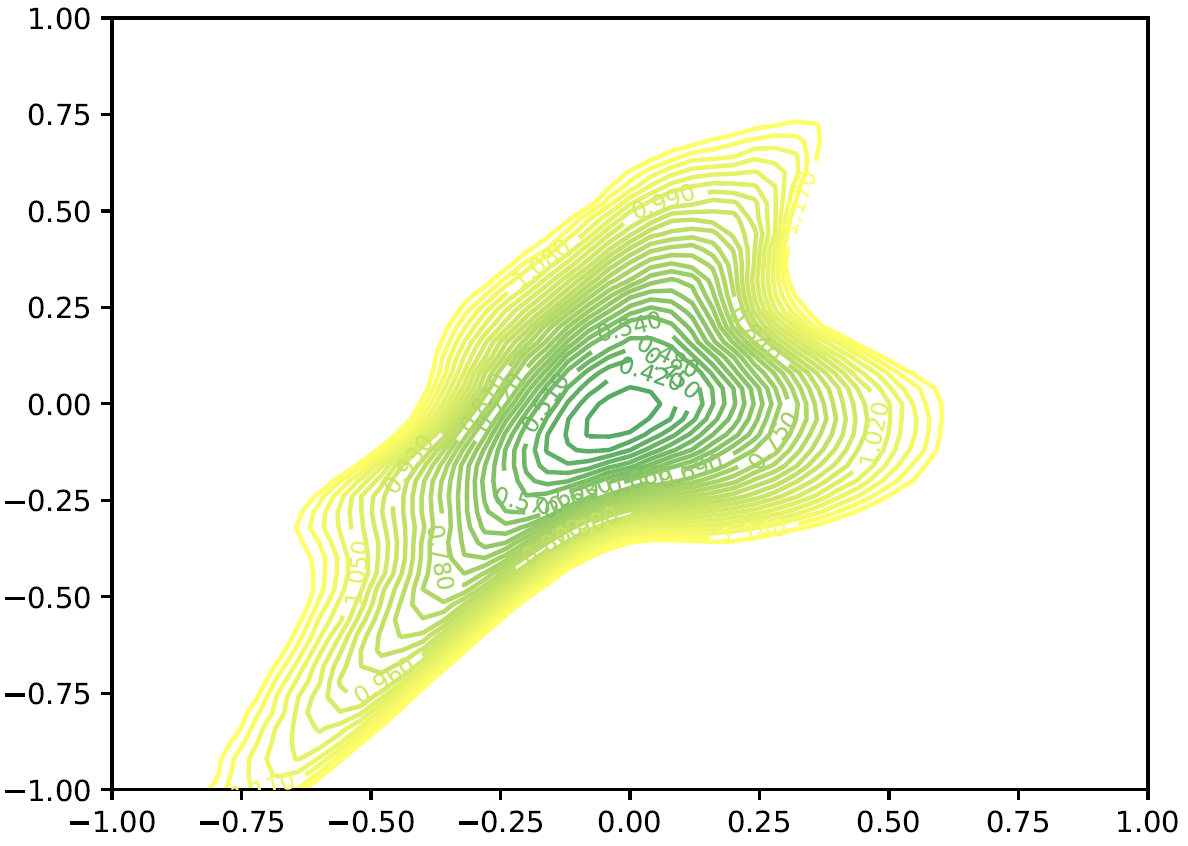}
         \caption{DARTS validation loss landscape.}
         \label{fig:fig:landscape-sharploss}
     \end{subfigure}
     \begin{subfigure}[b]{0.49\textwidth}
         \centering
         \includegraphics[width=\textwidth]{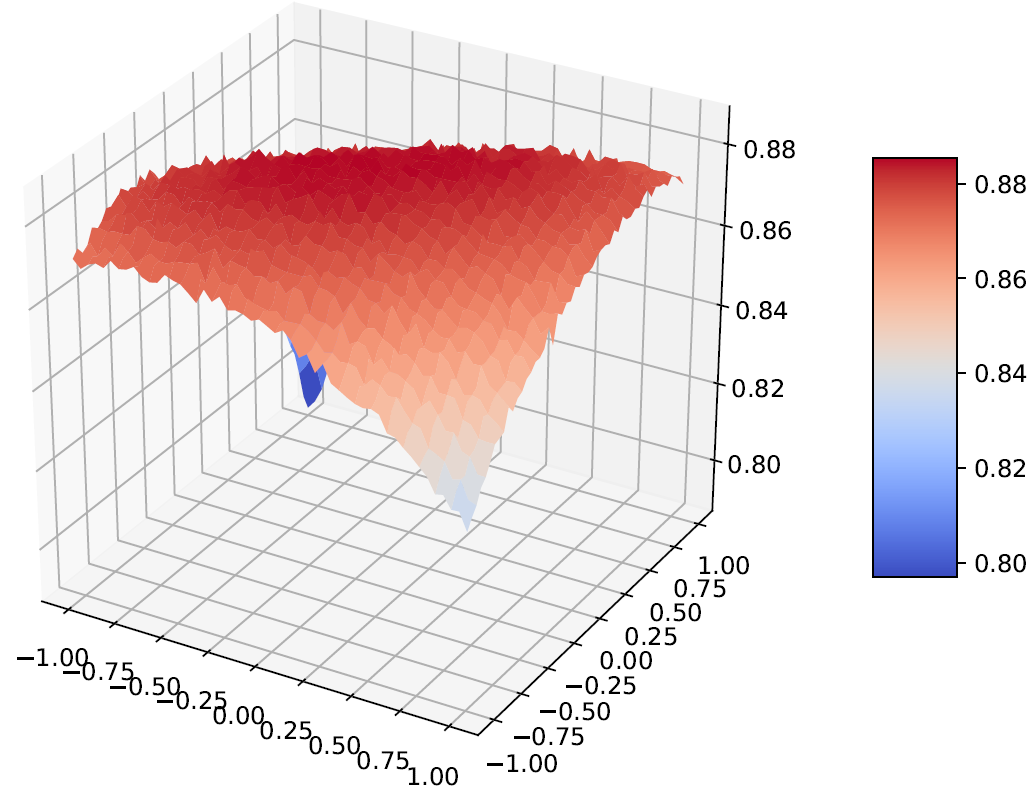}
         \caption{SA-DARTS validation accuracy landscape.}
         \label{fig:fig:landscape-flatacc}
     \end{subfigure}
     \begin{subfigure}[b]{0.49\textwidth}
         \centering
         \includegraphics[width=\textwidth]{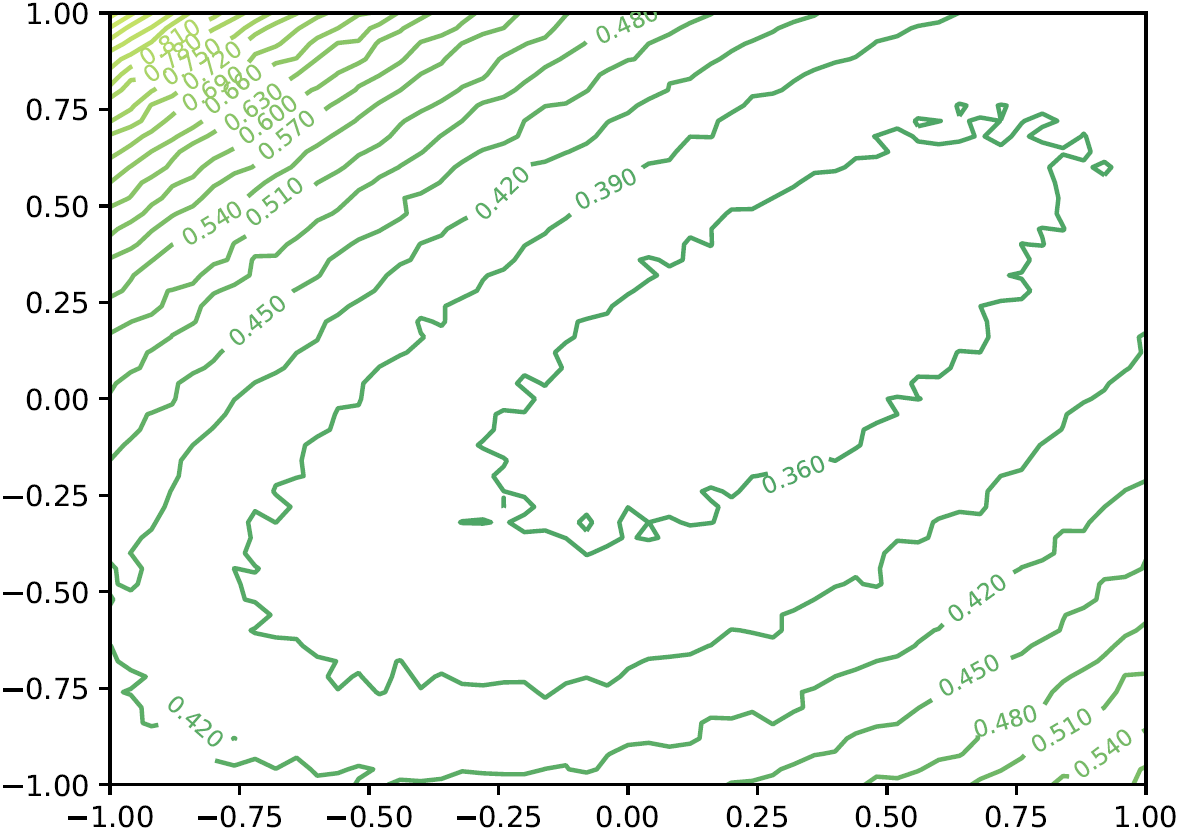}
         \caption{SA-DARTS validation loss landscape.}
         \label{fig:fig:landscape-flatloss}
     \end{subfigure}
        \caption{The visualization of validation accuracy and loss landscape with respect to architecture weights $\alpha$. Compared to original DARTS, our SA-DARTS smooths the landscape and stabilizes the searching process.}
        \label{fig:sharpflatcompare}
\end{figure*}

\section{Experimental Results}

This section demonstrates that our methods SA-DARTS and SAC-DARTS can constantly find the best architectures compared to other SOTA methods on classification and regression vision tasks.
In addition, our search method can find a better variant of existing SOTA models such as IMDN \cite{hui2019lightweight}.
For most tasks, the performance of the special case in Equation~\ref{eqn:mrloss} of regularizing mean values is not sensitive to the task type. 
We started with the special case mean regularizer in Equation~\ref{eqn:mrloss} for each task which can already give us SOTA. Then we tune $\nu$ in Equation~\ref{eqn:smu} to change the slope for negative inputs, finally, we can change smoothness by tuning $\mu$ in $erf$ function Equation~\ref{eqn:smu} for more improvements.  
Additional experiments including NAS-Bench-201, super-resolution with FLOPs constraint, negative initialization, and searched models are provided in Appendix.

\subsection{Validation Accuracy and Loss Landscape}

We compare the accuracy of loss landscape of SA-DARTS vs. original DARTS and show that SA-DARTS can land on a smoother validation loss/accuracy landscape.
Following the reduced search space of our previous simplified example, we define candidate operations per edge to be {3x3 SepConv, Skip Connect, Average Pooling} and search on CIFAR-10.
To visualize, we choose two random directions and apply normalized perturbation on $\alpha$, similar to \cite{li2018visualizing}.
As shown in Fig.~\ref{fig:sharpflatcompare}, the searched model of our SA-DARTS can land on a flatter validation accuracy and loss landscape compared to the original DARTS.


\subsection{Classification with DARTS space}

\begin{table*}[tpbh]
    \centering
    \resizebox{\textwidth}{!}{
    \begin{tabular}{|p{2.8cm}| p{1.2cm}|p{1.7cm}|p{1.2cm}|p{1.7cm}|p{1.3cm}|p{1cm}|p{1.3cm}|p{1.2cm}|}
\hline
        Method &  \multicolumn{2}{c}{Cifar 10} & \multicolumn{2}{c}{Cifar 100}  & \multicolumn{3}{c}{ImageNet} & Cost \\
        \hline
        & Params (M) & Accuracy (\%) & Params (M) & Accuracy (\%) & Params (M) & Top1 (\%) & Top5 (\%) & (Days)\\
\hline
        DARTS (1st) \cite{liu2018darts} &     3.4 & 97.00$\pm$0.14	& 3.4 & 82.46 & 4.9 & 73.3 & 91.3 & 0.4 \\
        DARTS- \cite{chu2020darts} $\square$ &      3.5 & 97.41$\pm$0.08 & 3.4 & 82.49$\pm$0.25 & 4.9 & 76.2 & 93.0 & 0.4 \\
        FairDARTS \cite{chu2020fair} $\square$ &    2.8 & 97.46$\pm$0.05 & -   & -     & 4.3 & 75.6 & 92.6 & 0.4 \\
        DrNAS  \cite{chen2020drnas}              &   4.0 & 97.46$\pm$0.03 & -   & -     & 5.2 & 75.8 & 92.7 & 3.9 \\     
        P-DARTS \cite{chen2019progressive} &        3.4 & 97.19$\pm$0.14 & 3.6 & 82.51 & 5.1 & 75.3 & 92.5 & 0.3 \\
        PC-DARTS \cite{xu2021partially} $\dagger$ &	3.6	& 97.43$\pm$0.07 & 3.6 & 83.10 & 5.3 & 75.8 & 92.7 & 0.1 \\
        SDARTS \cite{chen2020stabilizing} &               3.3 & 97.39$\pm$0.02 & -   & -     & 5.4 & 74.8 & 92.2 & 1.3 \\
        Beta-DARTS \cite{ye2022b} $\dagger$ &	3.8	& 97.47$\pm$0.08 & 3.8 & 83.76$\pm$0.22 & 5.5 & 76.1 & 93.0 & 0.4 \\
        Beta-DARTS $\diamond$ &	3.8	& 97.33$\pm$0.11 & 3.8 & 82.95$\pm$0.03 & 5.5 & 75.8 & 92.8 & 0.4 \\
        U-DARTS \cite{huang2023u}              &   3.3 & 97.41$\pm$0.06 & 3.4   &   83.54$\pm$0.06   & 4.9 & 73.9 & 91.9 & 4 \\  
        \hline
        PC + Beta & 3.8$\pm$0.07 & 97.42$\pm$0.05 & 3.8$\pm$0.07 & 82.95$\pm$0.07 & 5.4 & 75.9 & 92.6 & 0.1 \\
        SA-DARTS   & 4.1$\pm$0.11 & 97.40$\pm$0.06 & 4.1$\pm$0.12 & 83.73$\pm$0.08 & 5.5 & 75.7 & 92.6 & 0.4\\
        SAC-DARTS &  3.8$\pm$0.09 & 97.49$\pm$0.04 & 3.8$\pm$0.09 & 83.66$\pm$0.12 & 5.5 & 76.1 & 92.9 & 0.1 \\
    \hline
    \end{tabular}
    }
    \caption{Comparison of SOTA models on CIFAR10, CIFAR100, and ImageNet with DARTS Search Space. The CIFAR100 models incur about 0.05M additional parameters due to the classifier. $\dagger$ denotes the results reported in the paper. $\diamond$ denotes our reproduced results with DARTS codebase and training pipeline with \textit{random seeds}. $\square$ denotes that the model has a different search space. Reported DrNAS results are without progressive learning.}
    \label{tab:pcb-dartsspace}
\end{table*}

Following DARTS \cite{liu2018darts}, we search a classification network on CIFAR-10 with DARTS search space.
During the search phase, we define a bi-chain style supernet with repetitively stacked 6 normal cells and 2 reduction cells. 
The search settings are the same as DARTS since our method only introduces a simple regularization.
For evaluation, we follow DARTS for CIFAR10/CIFAR100 and PC-DARTS for ImageNet.
As shown in Table~\ref{tab:pcb-dartsspace}, our method can achieve SOTA performance on all three classification tasks. 
We generate our own baseline for fair comparison of PC-DARTS and Beta-DARTS on 10 \textit{random seeds}.
The search time is reduced from DARTS's 0.4 GPU days to 0.1 using SAC-DARTS.
For consistency, we jointly report the results for both reproduced Beta-DARTS and our PC-DARTS for CIFAR 100 and ImageNet.
We include this step because we also design \textbf{PC-Beta}, Beta-DARTS with partial channel trick, as a baseline.

SAC-DARTS can successfully reach SOTA, or even better results, with 1/4 the search time of SA-DARTS.
We also see a more steady validation result with SA-DARTS and SAC-DARTS.
For ImageNet, following other work, the best results of 10 \textit{random seeds} are reported.
SAC-DARTS clearly shows an advantage in both speed and accuracy, becoming the new SOTA for DARTS.


\subsection{Extension to Super-Resolution} \label{sec:sr}

\begin{figure}[tpbh]
     \centering
     \begin{subfigure}[b]{0.49\textwidth}
         \centering         \includegraphics[width=\textwidth]{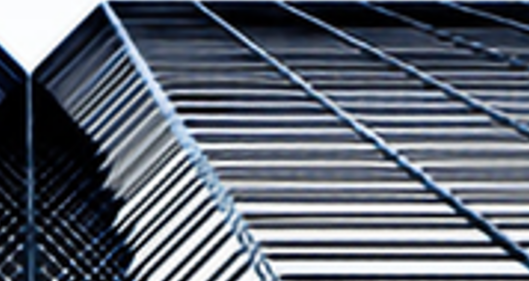}
         \caption{DLSR performance on Urban100, showing the issue with curvy strips.}
         \label{fig:dlsrurban}
     \end{subfigure}
     \hfill
     \begin{subfigure}[b]{0.49\textwidth}
         \centering         \includegraphics[width=\textwidth]{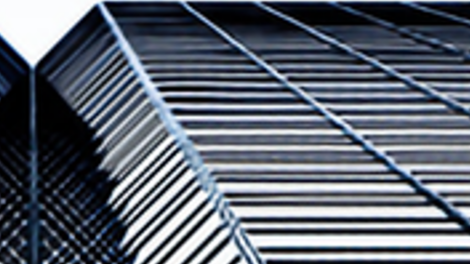}
         \caption{SAC-DARTS performance on Urban100, resolved the curvy strip issue.}
         \label{fig:mrurban}
     \end{subfigure}
        \caption{The visualization of DLSR vs. our SAC-DARTS on Urban100 Super-Resolution task (x2). The original DLSR is not capable of reconstructing the straight lines, but SAC-DARTS does not have this issue.}
        \label{fig:dlsrurbancompare}
\end{figure}

In this section, we switch our task to super-resolution (SR).
We consider two search spaces: DLSR \cite{huang2021lightweight} and information multi-distillation network (IMDN) \cite{hui2019lightweight}.
DLSR  uses the original DARTS algorithm to search for SR models.
We use our SA or SAC to improve DLSR.
All models are searched on DIV2K \cite{agustsson2017ntire} and evaluated on Set5, Set14, B100, and Urban100.
In Table~\ref{tab:psnr_dlsrspace}, we show the results on the reproduced DLSR, and the regularized versions as a comparison.
Again, we generate our own baselines here by following the same searching and training procedure in \cite{huang2021lightweight}, but without Flickr2K for a fair comparison with IMDN \cite{hui2019lightweight}. 
As shown in Table~\ref{tab:psnr_dlsrspace}, we give 2.2 GPU days search budget for all search methods, and 1.2 for SAC.
We also report the time when the searched architecture stops changing.
Our SA regularization-based methods require less time to converge to the final architecture.
SA-regularized DLSR only needs 0.4 days, and 0.11 for SAC to converge. 
Results are shown in Table~\ref{tab:psnr_dlsrspace} (Fig.~\ref{fig:dlsrurbancompare} for visual comparison).
Under the same searching and training pipeline, SA-DARTS and SAC-DARTS perform the best, achieving SOTA for all SR datasets.
Note that, Beta-regularization \cite{ye2022b} shows poor generalization ability on DLSR because it keeps alternating between two architectures, even after 2 days of searching.

\begin{table*}[t]
    \centering
    \resizebox{\textwidth}{!}{
    \begin{tabular}{|l|c|c|c|c|c|c|c|}
    \hline
       Method &		Search Cost (days) & Set5 &	Set14 &	B100 &	Urban100		& Params (K) &	Multi-Adds (G)\\
       \hline
DLSR &	 2+ &	37.91 &	33.56	&32.17 &	32.10 &	322.62 & 363.37	 \\
IMDN & - &	38.00 &	33.63&	32.19&	32.17&	715.18 & 654.53 \\
DLSR + Beta-DARTS &	 2.2 / 2.2  &	37.88&	33.46&	32.09&	31.71&	2243.00 &	2131.14\\
\hline
SA-DLSR (Ours) & 0.4 / 2.2  & 37.97 &	33.53 &	32.16 &	32.02 &	264.73	& 215.90 \\
SAC-DLSR (Ours) & 0.11 / 1.2 & 37.98 &	33.58 &	32.17 &	32.16 &	309.37 &	350.70 \\
\hline
IMDN + Beta-DARTS &	 2.2 / 2.2  &	\textbf{38.01} &	33.47&	32.11&	32.12&	1773.67 &	1745.04 \\
SAC-IMDN (Ours) & 0.7/1.2  & 37.98 &	\textbf{33.65} &	\textbf{32.20} &	\textbf{32.21} &	349.22	& 317.09 \\
    \hline
    \end{tabular}
    }
    \caption{PSNR (x2 scale) results of different methods. Search is on DIV2K only, and evaluated on different SR Datasets with DLSR Search Space. The model searched by SA-DARTS achieves the best PSNR value among all methods with the DLSR search space, and can match IMDN \cite{hui2019lightweight} (different model space) with only half of the parameters. First number under Search Cost is the time spent when architecture stops changing.}
    \label{tab:psnr_dlsrspace}
\end{table*}

\subsection{Model Optimization with SA-DARTS}

In addition, we use SAC-DARTS to find better variants of the original IMDN model. 
The IMDN has 6 repetitive IMD blocks, and each block has 4 conv3x3 operators.
We keep the overall layer design unchanged and optimize conv3x3 operators in IMD blocks. 
More details in Fig.~\ref{fig:imdn-nas}.
There are 12 candidates to potentially replace each conv3x3 operator.
There are three convolution types: normal conv, depth separable conv, and dilated depth sep conv, with kernel choices of 3x3 or 5x5, and with or without batch normalization.
For both DLSR and IMDN search space, Beta-DARTS searched models are a lot larger than SA or SAC, but the performances are not any better.
For DLSR, Beta-DARTS searched conv7x7 for all 3 blocks, while for IMDN search space, Beta-DARTS found conv5x5 for all blocks, all 4 blocks have batch norm except for the second.

\subsubsection{IMDN Search Space}

As shown in 
For IMDN search space, SA-DARTS searches 4 operators in purple and try to replace the original conv3x3 by other operators.
The candidates are: conv, depth separable conv, dilated depth separable conv with kernal choice of 3x3 or 5x5, and choice of followed by batch norm. 
In total, 12 candidates are given.
The improved IMDN model found by SA-DARTS can be found in Appendix Fig.~\ref{fig:imdn-nas}.
Without FLOPs constraint, Beta-DARTS picked all conv5x5 operators, Fig.~\ref{fig:imdn-nas}(a) shows final model searched by SA-DARTS. 
For SAC-DARTS, its genotype is: \textit{[dilated depth sep conv3x3, depth sep conv3x3, depth sep conv3x3, dilated depth sep conv3x3].}


\section{Conclusion}

In this work, we have proposed a regularization technique that is unique to the DARTS searching task.
Updating architecture weight $\alpha$ is very different from traditional optimization of DNN weights, thus L1, L2, or weight-decay are less effective.
Our method SA-DARTS adds an additional loss term by encapsulating $\alpha$ inside a smooth activation function to regularize the update.
SA-DARTS is designed to control the relative ranking among all candidates better.
We show that SA-DARTS and SAC-DARTS eliminate the skip dominance, can recover from an unfair local optimal, and mitigate the discrepancy of discretization with widespread $\alpha$ values.
We demonstrate the efficacy of proposed algorithms through different tasks and search spaces, \textit{e.g.,}, classification with DARTS and NAS-Bench-201 search space, super-resolution with DLSR and IMDN space.


\bibliographystyle{plain}
\bibliography{sadarts}

\begin{thebibliography}{10}

\bibitem{agustsson2017ntire}
Eirikur Agustsson and Radu Timofte.
\newblock Ntire 2017 challenge on single image super-resolution: Dataset and study.
\newblock In {\em Proceedings of the IEEE conference on computer vision and pattern recognition workshops}, pages 126--135, 2017.

\bibitem{biswas2021smu}
Koushik Biswas, Sandeep Kumar, Shilpak Banerjee, and Ashish~Kumar Pandey.
\newblock Smu: Smooth activation function for deep networks using smoothing maximum technique.
\newblock {\em arXiv preprint arXiv:2111.04682}, 2021.

\bibitem{chen2020stabilizing}
Xiangning Chen and Cho-Jui Hsieh.
\newblock Stabilizing differentiable architecture search via perturbation-based regularization.
\newblock In {\em International conference on machine learning}, pages 1554--1565. PMLR, 2020.

\bibitem{chen2020drnas}
Xiangning Chen, Ruochen Wang, Minhao Cheng, Xiaocheng Tang, and Cho-Jui Hsieh.
\newblock Drnas: Dirichlet neural architecture search.
\newblock {\em arXiv preprint arXiv:2006.10355}, 2020.

\bibitem{chen2019progressive}
Xin Chen, Lingxi Xie, Jun Wu, and Qi~Tian.
\newblock Progressive differentiable architecture search: Bridging the depth gap between search and evaluation.
\newblock In {\em Proceedings of the IEEE/CVF International Conference on Computer Vision}, pages 1294--1303, 2019.

\bibitem{chen2020spatiotemporal}
Zhifeng Chen, Hantao Wang, Lijun Wu, Yanlin Zhou, and Dapeng Wu.
\newblock Spatiotemporal guided self-supervised depth completion from lidar and monocular camera.
\newblock In {\em 2020 IEEE International conference on visual communications and image processing (VCIP)}, pages 54--57. IEEE, 2020.

\bibitem{chu2020darts}
Xiangxiang Chu, Xiaoxing Wang, Bo~Zhang, Shun Lu, Xiaolin Wei, and Junchi Yan.
\newblock Darts-: robustly stepping out of performance collapse without indicators.
\newblock {\em arXiv preprint arXiv:2009.01027}, 2020.

\bibitem{chu2021fairnas}
Xiangxiang Chu, Bo~Zhang, and Ruijun Xu.
\newblock Fairnas: Rethinking evaluation fairness of weight sharing neural architecture search.
\newblock In {\em Proceedings of the IEEE/CVF International Conference on Computer Vision}, pages 12239--12248, 2021.

\bibitem{chu2020fair}
Xiangxiang Chu, Tianbao Zhou, Bo~Zhang, and Jixiang Li.
\newblock Fair darts: Eliminating unfair advantages in differentiable architecture search.
\newblock In {\em European conference on computer vision}, pages 465--480. Springer, 2020.

\bibitem{deng2023distributionally}
Menghua Deng, Bomin Bian, Yanlin Zhou, and Jianpeng Ding.
\newblock Distributionally robust production and replenishment problem for hydrogen supply chains.
\newblock {\em Transportation Research Part E: Logistics and Transportation Review}, 179:103293, 2023.

\bibitem{deng2024stochastic}
Menghua Deng, Yuanbo Li, Jianpeng Ding, Yanlin Zhou, and Lianming Zhang.
\newblock Stochastic and robust truck-and-drone routing problems with deadlines: A benders decomposition approach.
\newblock {\em Transportation Research Part E: Logistics and Transportation Review}, 190:103709, 2024.

\bibitem{el2025attentive}
Mostafa El-Khamy, HOR Soheil, and Yanlin Zhou.
\newblock Attentive sensing for efficient multimodal gesture recognition, March~13 2025.
\newblock US Patent App. 18/882,626.

\bibitem{el2024systems}
Mostafa El-Khamy and Yanlin Zhou.
\newblock Systems and methods for neural architecture search, February~29 2024.
\newblock US Patent App. 18/148,418.

\bibitem{guo2021unsupervised}
Ente Guo, Zhifeng Chen, Yanlin Zhou, and Dapeng~Oliver Wu.
\newblock Unsupervised learning of depth and camera pose with feature map warping.
\newblock {\em Sensors}, 21(3):923, 2021.

\bibitem{hor2024cm}
Soheil Hor, Mostafa El-Khamy, Yanlin Zhou, Amin Arbabian, and SukHwan Lim.
\newblock Cm-asap: Cross-modality adaptive sensing and perception for efficient hand gesture recognition.
\newblock In {\em 2024 IEEE 7th International Conference on Multimedia Information Processing and Retrieval (MIPR)}, pages 207--213. IEEE, 2024.

\bibitem{huang2021lightweight}
Han Huang, Li~Shen, Chaoyang He, Weisheng Dong, Haozhi Huang, and Guangming Shi.
\newblock Lightweight image super-resolution with hierarchical and differentiable neural architecture search.
\newblock {\em arXiv preprint arXiv:2105.03939}, 2021.

\bibitem{huang2023u}
Lan Huang, Shiqi Sun, Jia Zeng, Wencong Wang, Wei Pang, and Kangping Wang.
\newblock U-darts: Uniform-space differentiable architecture search.
\newblock {\em Information Sciences}, 628:339--349, 2023.

\bibitem{hui2019lightweight}
Zheng Hui, Xinbo Gao, Yunchu Yang, and Xiumei Wang.
\newblock Lightweight image super-resolution with information multi-distillation network.
\newblock In {\em Proceedings of the 27th acm international conference on multimedia}, pages 2024--2032, 2019.

\bibitem{kwon2024method}
Hyukjoon Kwon, Mohamed Mahmoud, Federico Penna, Yanlin Zhou, and Ramy~E Ali.
\newblock Method and apparatus for learning-based channel matrix prediction, September~10 2024.
\newblock US Patent 12,088,369.

\bibitem{li2018visualizing}
Hao Li, Zheng Xu, Gavin Taylor, Christoph Studer, and Tom Goldstein.
\newblock Visualizing the loss landscape of neural nets.
\newblock {\em Advances in neural information processing systems}, 31, 2018.

\bibitem{liang2019darts+}
Hanwen Liang, Shifeng Zhang, Jiacheng Sun, Xingqiu He, Weiran Huang, Kechen Zhuang, and Zhenguo Li.
\newblock Darts+: Improved differentiable architecture search with early stopping.
\newblock {\em arXiv preprint arXiv:1909.06035}, 2019.

\bibitem{liu2018darts}
Hanxiao Liu, Karen Simonyan, and Yiming Yang.
\newblock Darts: Differentiable architecture search.
\newblock {\em arXiv preprint arXiv:1806.09055}, 2018.

\bibitem{ma2019retailnet}
Xiyao Ma, Fan Lu, Xiajun~Amy Pan, Yanlin Zhou, and Xiaolin~Andy Li.
\newblock Retailnet: Enhancing retails of perishable products with multiple selling strategies via pair-wise multi-q learning.
\newblock 2019.

\bibitem{ma2020improving}
Xiyao Ma, Qile Zhu, Yanlin Zhou, and Xiaolin Li.
\newblock Improving question generation with sentence-level semantic matching and answer position inferring.
\newblock In {\em Proceedings of the AAAI conference on artificial intelligence}, volume~34, pages 8464--8471, 2020.

\bibitem{ma2020asking}
Xiyao Ma, Qile Zhu, Yanlin Zhou, Xiaolin Li, and Dapeng Wu.
\newblock Asking complex questions with multi-hop answer-focused reasoning.
\newblock {\em arXiv preprint arXiv:2009.07402}, 2020.

\bibitem{peng2017distributed}
Chen Peng, Yanlin Zhou, and Qing Hui.
\newblock Distributed fault diagnosis with shared-basis and b-splines-based matched learning.
\newblock In {\em 2017 13th IEEE Conference on Automation Science and Engineering (CASE)}, pages 536--541. IEEE, 2017.

\bibitem{peng2019distributed}
Chen Peng, Yanlin Zhou, and Qing Hui.
\newblock Distributed fault diagnosis of networked dynamical systems with time-varying topology.
\newblock {\em Journal of the Franklin Institute}, 356(11):5754--5780, 2019.

\bibitem{pu2021server}
George Pu, Yanlin Zhou, Dapeng Wu, and Xiaolin Li.
\newblock Server averaging for federated learning.
\newblock {\em arXiv preprint arXiv:2103.11619}, 2021.

\bibitem{wang2021rethinking}
Ruochen Wang, Minhao Cheng, Xiangning Chen, Xiaocheng Tang, and Cho-Jui Hsieh.
\newblock Rethinking architecture selection in differentiable nas.
\newblock {\em arXiv preprint arXiv:2108.04392}, 2021.

\bibitem{xie2018snas}
Sirui Xie, Hehui Zheng, Chunxiao Liu, and Liang Lin.
\newblock Snas: stochastic neural architecture search.
\newblock {\em arXiv preprint arXiv:1812.09926}, 2018.

\bibitem{xu2021partially}
Yuhui Xu, Lingxi Xie, Wenrui Dai, Xiaopeng Zhang, Xin Chen, Guo-Jun Qi, Hongkai Xiong, and Qi~Tian.
\newblock Partially-connected neural architecture search for reduced computational redundancy.
\newblock {\em IEEE Transactions on Pattern Analysis and Machine Intelligence}, 43(9):2953--2970, 2021.

\bibitem{xu2019pc}
Yuhui Xu, Lingxi Xie, Xiaopeng Zhang, Xin Chen, Guo-Jun Qi, Qi~Tian, and Hongkai Xiong.
\newblock Pc-darts: Partial channel connections for memory-efficient architecture search.
\newblock {\em arXiv preprint arXiv:1907.05737}, 2019.

\bibitem{ye2022b}
Peng Ye, Baopu Li, Yikang Li, Tao Chen, Jiayuan Fan, and Wanli Ouyang.
\newblock b-darts: Beta-decay regularization for differentiable architecture search.
\newblock In {\em Proceedings of the IEEE/CVF Conference on Computer Vision and Pattern Recognition}, pages 10874--10883, 2022.

\bibitem{zela2019understanding}
Arber Zela, Thomas Elsken, Tonmoy Saikia, Yassine Marrakchi, Thomas Brox, and Frank Hutter.
\newblock Understanding and robustifying differentiable architecture search.
\newblock {\em arXiv preprint arXiv:1909.09656}, 2019.

\bibitem{zhou2020theory}
Pan Zhou, Caiming Xiong, Richard Socher, and Steven Chu~Hong Hoi.
\newblock Theory-inspired path-regularized differential network architecture search.
\newblock {\em Advances in Neural Information Processing Systems}, 33:8296--8307, 2020.

\bibitem{zhou2018design}
Yanlin Zhou.
\newblock Design of a distributed real-time e-health cyber ecosystem with collective actions: Diagnosis, dynamic queueing, and decision making.
\newblock 2018.

\bibitem{zhou2018developing}
Yanlin Zhou, Ryan Anderson, Hamid Vakilzadian, Dietmar~PF Moeller, and Andreas Deutschmann.
\newblock Developing a dynamic queueing model for the airport check-in process.
\newblock In {\em 2018 IEEE International Conference on Electro/Information Technology (EIT)}, pages 0871--0876. IEEE, 2018.

\bibitem{zhou2019adaptive}
Yanlin Zhou, Fan Lu, George Pu, Xiyao Ma, Runhan Sun, Hsi-Yuan Chen, and Xiaolin Li.
\newblock Adaptive leader-follower formation control and obstacle avoidance via deep reinforcement learning.
\newblock In {\em 2019 IEEE/RSJ International Conference on Intelligent Robots and Systems (IROS)}, pages 4273--4280. IEEE, 2019.

\bibitem{zhou2022communication}
Yanlin Zhou, Xiyao Ma, Dapeng Wu, and Xiaolin Li.
\newblock Communication-efficient and attack-resistant federated edge learning with dataset distillation.
\newblock {\em IEEE Transactions on Cloud Computing}, 11(3):2517--2528, 2022.

\bibitem{zhou2018spiking}
Yanlin Zhou, Chen Peng, and Qing Hui.
\newblock A spiking neural dynamical drift-diffusion model on collective decision making with self-organized criticality.
\newblock In {\em 2018 Annual American Control Conference (ACC)}, pages 5645--5652. IEEE, 2018.

\bibitem{zhoumodular}
Yanlin Zhou, George Pu, Fan Lu, Xiyao Ma, and Xiaolin Li.
\newblock Modular platooning and formation control.

\bibitem{zhou2020distilled}
Yanlin Zhou, George Pu, Xiyao Ma, Xiaolin Li, and Dapeng Wu.
\newblock Distilled one-shot federated learning.
\newblock {\em arXiv preprint arXiv:2009.07999}, 2020.

\end{thebibliography}

\newpage

\section{Appendix}

\subsection{Best Architecture}

In this section, we present our best architectures searched on CIFAR100.

\begin{figure}[tpbh]
    \centering
    \includegraphics[width=0.45\textwidth]{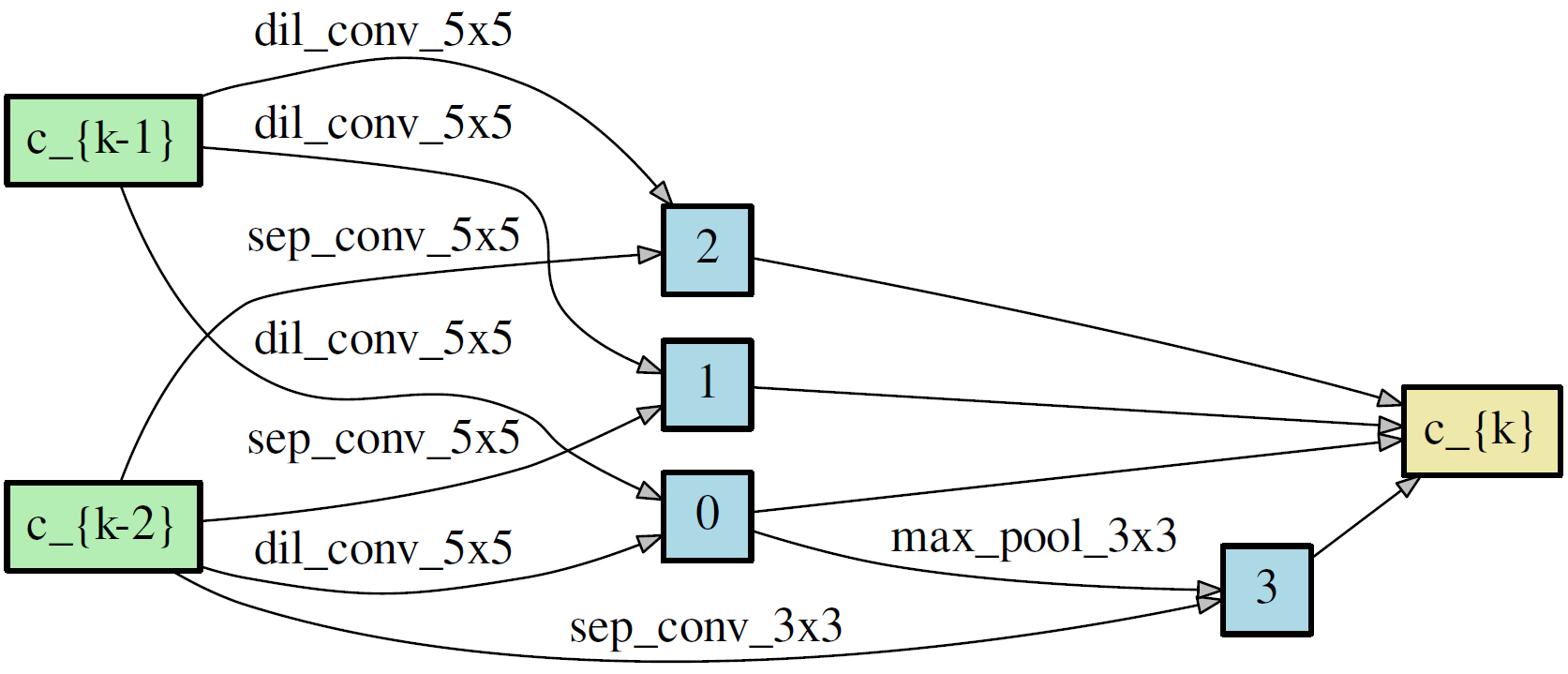}
    \caption{Current best normal cell of SAC-DARTS with first 15 epochs on weights only.}
    \label{fig:mr-normal}
\end{figure}

\begin{figure}[tpbh]
    \centering
    \includegraphics[width=0.45\textwidth]{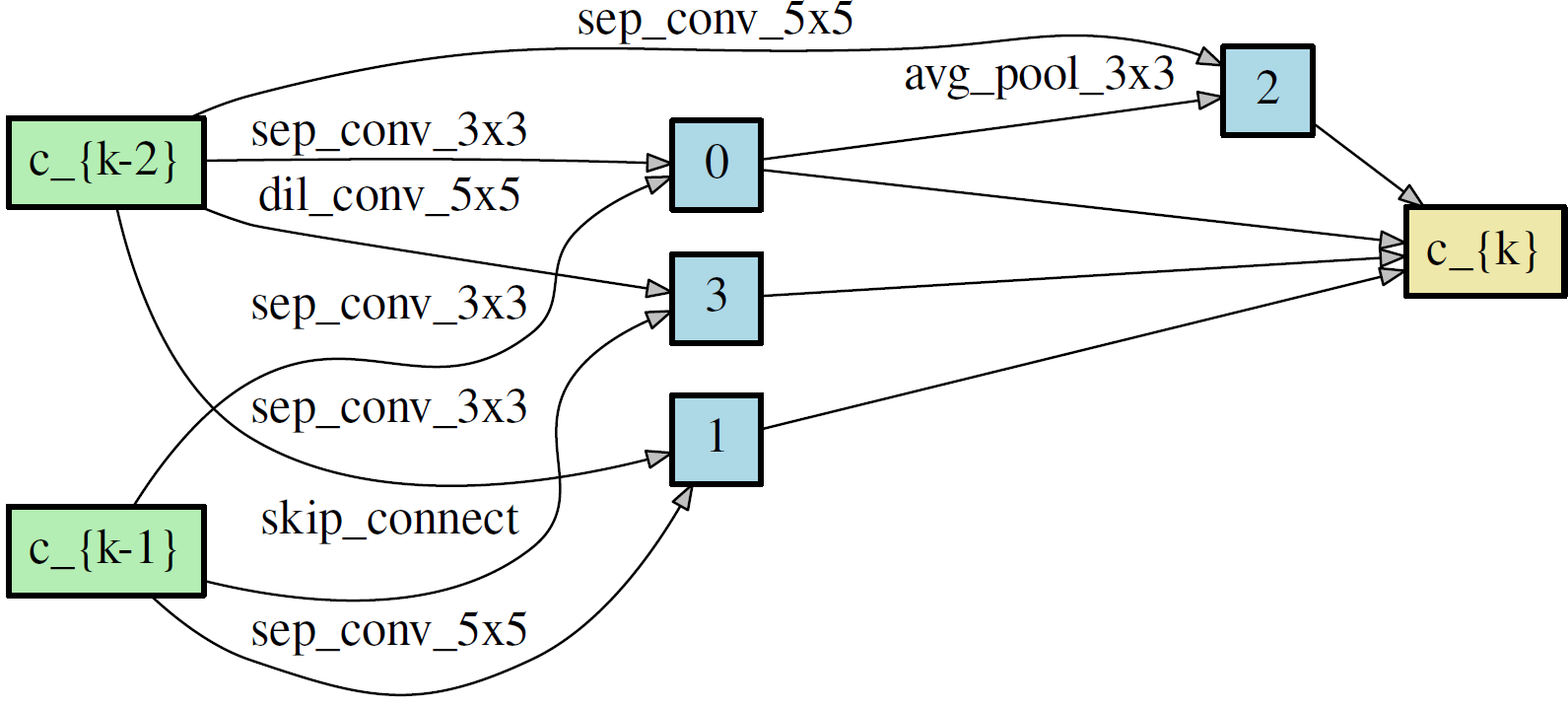}
    \caption{Current best reduction cell of SAC-DARTS with first 15 epochs on weights only.}
    \label{fig:mr-reduced}
\end{figure}


\subsection{Implementation Details}

All experiments are run on a single V100 GPU.
For NAS-Bench-201, half of the dataset is used for training, and the batch size is set to 64. 
$\alpha$ is initialized to be small random numbers close to $\num{1E-3}$.
To update $\alpha$ , Adam is used with $\num{3E-4}$ learning rate, as well as $\num{1E-3}$ weight decay.
Note that weight decay is removed for SA regularization, but turned on for all other reproduced results following their public code.
A cosine Annealing scheduler is used.
To update network weights, SGD optimizer with $0.9$ momentum is used with $\num{2.5E-2}$ learning rate, as well as $\num{3e-4}$ weight decay.
The initial channel is set to 16.
To combat overfitting and to further improve results, the gradient clip is set to 5.0, and cutout is used.
50 epochs are scheduled for all methods, but our SA-DARTS and SAC-DARTS converge within 10 epochs in most cases without warmup.
Note that warmup refers to updating only the network weight and freezing the alpha weights in the first 15 epochs.
This trick can help combat the skip dominance issue and is usually used in many other literatures, such as PC-DARTS, P-DARTS, and DLSR.
Our SA-DARTS and SAC-DARTS can converge to better solutions without warmup, and further improve the convergence speed of SOTA works.

For classification tasks, we define a bi-chain style supernet for \textbf{DARTS search space} with repetitively stacked 6 normal cells and 2 reduction cells. 
Each cell has 4 intermediate nodes with 14 edges and each edge can choose from 8 candidate operations.
When searching on CIFAR10 or CIFAR100, half of the dataset is used for training and the batch size is set to 256.
$\alpha$ is initialized to be small random numbers close to $\num{1E-3}$.
To update $\alpha$ , adam is used with $\num{6E-4}$ learning rate, as well as $\num{1E-3}$ weight decay.
Note that weight decay is removed for SA regularization, but turned on for all other reproduced results following their public code.
Cosine Annealing scheduler is used.
To update network weight, SGD optimizer with $0.9$ momentum is used with $0.1$ learning rate, as well as $\num{3e-4}$ weight decay.
The initial channel is set to 16.
The gradient clip is set to 5.0 and cutout is used.
While searching only uses 8 cells, evaluation uses 20 cells trained from scratch.
Models searched on CIFAR10 and CIFAR100 are both transferred to the ImageNet model following DARTS \cite{liu2018darts}.

For \textbf{super resolution} tasks, we follow DLSR \cite{huang2021lightweight} for hyperparameters.
Adam optimizer is used for both network and architecture weights.
For network weights, learning rate is $\num{5E-4}$, weight decay is $\num{1E-8}$.
For architecture weights, learning rate is $\num{3E-4}$, weight decay is $\num{1E-8}$.
For fair comparison with IMDN, the model is searched on DIV2K \cite{agustsson2017ntire}.
Parameters and FLOPs are all calculated on images with 1280x720 resolution.


\textbf{Reduced Search Space}
The authors in \cite{zela2019understanding} discovered DARTS's skip dominance issue in 4 reduced search spaces. 
We illustrate the comparison on 4 simplified search spaces containing a portion of candidate operations in Table~\ref{tab:reduced_space}.
We use 20 cells with 36 initial channels for CIFAR-10. 
Our methods achieve consistent performance gain compared with baselines.

\begin{table}[h]
    \centering
    \begin{tabular}{|l|c|c|c|c|}
    \hline
        Method & S1 &  S2 & S3 & S4 \\
    \hline
        DARTS  \cite{liu2018darts} & 3.84 & 4.85 & 3.34 & 7.20  \\
        R-DARTS \cite{ye2022b} & 3.11 & 3.48 & 2.93 & 3.58 \\
        SA-DARTS (ours)   &  2.78 & 3.02 & 2.49 & 2.87\\
    \hline
    \end{tabular}
    \caption{Test error (\%) on CIFAR10 of different models on reduced DARTS search space S3.}
    \label{tab:reduced_space}
\end{table}


\begin{table*}[tpbh]
    \centering
    \resizebox{\textwidth}{!}{
    \begin{tabular}{|l|l|l|l|l|}
\hline
        Method & CIFAR10 Accuracy (\%) & CIFAR10 Accuracy (\%) & NasBench201 CIFAR10 (\%) & SR Set14 PSNR\\
\hline
SA-DARTS ($\nu=1$)   & 97.40 & 83.73 & 91.55 & 33.46  \\
SAC-DARTS ($\nu=1$)  & 97.49 & 83.66 & 91.55 &  33.47 \\
\hline
SA-DARTS ($\mu=1e6$,$\nu=0.25$)   & 97.41 & 83.71 & 91.55 & 33.50 \\
SAC-DARTS ($\mu=1e6$,$\nu=0.25$)  & 97.37 & 83.55 & 91.55 & 33.47 \\
\hline
SA-DARTS ($\mu=1\sqrt2$,$\nu=0$)   & 97.17 & 82.53 & 91.21 & 33.53 \\
SAC-DARTS ($\mu=1\sqrt2$,$\nu=0$)  & 97.04 & 82.97 & 91.21 & 33.58 \\
\hline
    \end{tabular}
    }
    \caption{Comparison of different hyperparameters $\mu$ and $\nu$, all results are best of 3. There exist many optimal sets of parameters to achieve SOTA results. Although the special case in Equation (8) can always deliver good result, some parameters may be better for different tasks.}
    \label{tab:mu_vu}
\end{table*}

\begin{figure}[tpbh]
    \centering
    \includegraphics[width=0.7\textwidth]{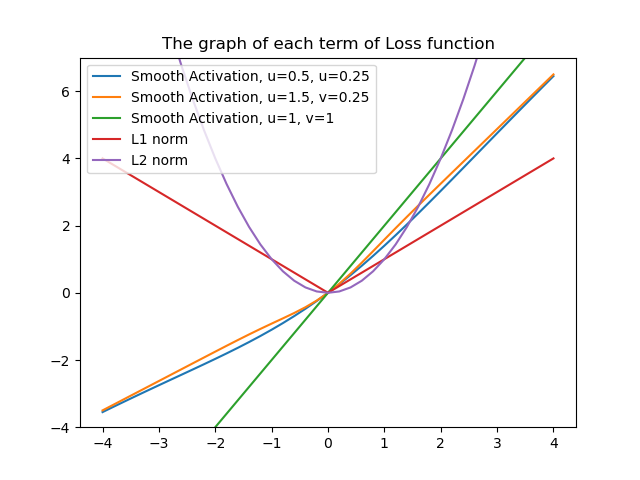}
    \caption{Different regularization methods. If we change $\mu$ in erf function, a smooth curvature is introduced to the origin}
    \label{fig:enter-label}
\end{figure}


\subsection{More on hyperparameters on Equation~\ref{eqn:smu}}
For most tasks, the performance of the special case in Eqn. \ref{eqn:mrloss} of regularizing mean values is not sensitive to the task type. In this work, we started with the special case mean regularizer in Eqn. \ref{eqn:mrloss} for each task which can already give us SOTA. Then we tune $vu$ in Eqn. \ref{eqn:smu} to change the slope for negative inputs, finally, we can change smoothness by tuning $mu$ in $erf$ function Eqn. \ref{eqn:smu} for more improvements.
In addition, once the $mu$ and $vu$ combination Eqn. \ref{eqn:smu} is in the optimal neighborhood, the performance is not sensitive to the change in $mu$ and $vu$. Through ablation study, we find many pairs of hyperparameters that can achieve SOTA results on NAS-Bench-201. For instance, when we set =1e6 (mimicking infinity), =0.25 or 0.75 will both achieve SOTA architecture in NASBench201 with CIFAR10 valid 91.55\%.


\subsection{More analysis on Equation~\ref{eqn:alpha_conv}}\label{sec:app_skip_dom}

The objective can be formulated as: 
$\text{min}_{\beta_{conv}, \beta_{skip}, \beta_{avg}} Var(\overline{m} (x) - m^*)$, such that $\beta_{conv} + \beta_{skip} + \beta_{avg} = 1$.

This constraint optimization problem can be solved with Lagrangian multipliers:
\begin{multline}
    {\cal L}(\beta_{conv}, \beta_{skip}, \beta_{avg},\lambda)=Var(\overline{m}(x)-m^*)- \\ \lambda(\beta_{conv} + \beta_{skip} + \beta_{avg}-1)\\
    =Var(\beta_{conv}o_c(x) + \beta_{skip} x + \beta_{avg} o_a(x)-m^*)- \\ \lambda(\beta_{conv} + \beta_{skip} + \beta_{avg}-1)\\
     =Var(\beta_{conv}(o_c(x)-m^*) + \beta_{skip}(x-m^*) + \beta_{avg} (o_a(x)-m^*)) \\ -\lambda(\beta_{conv} + \beta_{skip} + \beta_{avg}-1)\\
\end{multline}

Expanding Variance term and introducing covariance:
\begin{multline}
    =Var(\beta_{conv}(o_c(x)-m^*) + Var(\beta_{skip}(x-m^*)) + \\ Var(\beta_{avg} (o_a(x)-m^*)) + \\
    2Cov(\beta_{skip}(x-m^*),\beta_{avg} (o_a(x)-m^*))+\\
    2Cov(\beta_{conv}(o_c(x)-m^*),\beta_{avg} (o_a(x)-m^*))+\\
    2Cov(\beta_{conv}(o_c(x)-m^*),\beta_{skip}(x-m^*)) - \\ \lambda(\beta_{conv} +     \beta_{skip} + \beta_{avg}-1)\\  
\end{multline}

\begin{multline}
    =\beta_{conv}^2Var(o_c(x)-m^*) + \beta_{skip}^2Var(x-m^*) + \\ \beta_{avg}^2 Var(o_a(x)-m^*) + \\
    2\beta_{skip}\beta_{avg}Cov((x-m^*), (o_a(x)-m^*))+\\
    2\beta_{conv}\beta_{avg}Cov((o_c(x)-m^*), (o_a(x)-m^*))+\\
    2\beta_{conv}\beta_{skip}Cov((o_c(x)-m^*),(x-m^*)) - \\ \lambda(\beta_{conv} +     \beta_{skip} + \beta_{avg}-1)\\  
\end{multline}

Setting:
$\frac{\partial L}{\partial\lambda}, \frac{\partial L}{\partial\beta_{c o n v}}, \frac{\partial L}{\partial\beta_{s k i p}}, \frac{\partial L}{\partial\beta_{avg}}$ all to be 0.

\begin{multline}
    \frac{\partial L}{\partial\lambda}=\beta_{conv} +     \beta_{skip} + \beta_{avg}-1=0\\
    \frac{\partial L}{\partial\beta_{c o n v}}=2\beta_{c o n v}V a r(o_{c}(x)-m^{*})+ \\
2\beta_{avg}Cov((o_c(x)-m^*), (o_a(x)-m^*)) + \\
2\beta_{skip}Cov((o_c(x)-m^*),(x-m^*)) -\lambda = 0
\end{multline}

\begin{multline}
    \frac{\partial L}{\partial\beta_{skip}}=2\beta_{skip}V a r(x-m^{*})+\\
2\beta_{avg}Cov((x-m^*), (o_a(x)-m^*)) + \\
2\beta_{conv}Cov((o_c(x)-m^*),(x-m^*)) -\lambda = 0
\end{multline}

\begin{multline}
    \frac{\partial L}{\partial\beta_{avg}}=2\beta_{avg}V a r(o_{a}(x)-m^{*})+\\
2\beta_{skip}Cov((x-m^*), (o_a(x)-m^*)) + \\
2\beta_{conv}Cov((o_c(x)-m^*),(o_a(x)-m^*)) -\lambda = 0
\end{multline}

Following the 2 operator case in \cite{wang2021rethinking}, we can solve the above equations by recursion.

From the Equation~\ref{eqn:alphaconv} - \ref{eqn:alphaavg}, it is clear that the relative magnitudes of $\alpha_{conv}^*$, $\alpha_{skip}^*$, and $\alpha_{skip}^*$ greatly depend on whether an operation being applied takes it closer to the optimal feature map. 
While $x$ comes from the mixed output of the previous edge, it will naturally be closer to $m^*$, $O_a(x)$ is an average pooling, which will be slightly off but very consistent, and $O_c(x)$ is the output of a single convolution operation, which is inconsistent during training, becoming the furthest.

Following the simplified search space in Section~\ref{sec:peformancecollapse}, it can be easily inferred that $\alpha_{skip}$, $\alpha_{conv}$, and $\alpha_{avg}$:

\begin{multline}\label{eqn:alpha_convapp}
    \exp(\alpha_{conv}) = \text{Var}(x-\overline{m} (x))+\text{Var}(o_a(x)-\overline{m} (x)) - \\ \text{Cov}(x-\overline{m} (x), o_c(x)-\overline{m} (x), o_a(x)-\overline{m} (x)) + C
\end{multline}

\begin{multline}\label{eqn:alpha_skip}
    \exp(\alpha_{skip}) = \text{Var}(o_c(x)-\overline{m} (x))+  \text{Var}(o_a(x)-\overline{m} (x)) - \\  \text{Cov}(x-m^*, o_c(x)-m^*, o_a(x)-m^*) + C
\end{multline}

\begin{multline}\label{eqn:alpha_avg}
    \exp(\alpha_{avg}) = \text{Var}(x-\overline{m} (x))+\text{Var}(o_c(x)-\overline{m} (x)) - \\ \text{Cov}(x-\overline{m} (x), o_c(x)-\overline{m} (x), o_a(x)-\overline{m} (x)) + C
\end{multline}

Therefore, to reduce the value of $\text{Var}(o_c(x)-\overline{m} (x))+\text{Var}(x-\overline{m} (x))+\text{Var}(o_a(x)-\overline{m} (x))$, we can choose a smaller value for $\alpha$, such that $\exp(\alpha)$ is small.


\subsection{More analysis on Equation~\ref{eqn:smu}}

Through ablation study, we find many pairs of hyperparameters that can achieve SOTA results on NAS-Bench-201.
In addition to the mean regularization, another example is $\nu = 0.25$ and $\mu = $ \num{1e6}, almost infinity.
Other hyperparameters such as $\lambda = \text{epoch}/2$, $\nu = 0.25$ and $\mu = 0$ degrade the test accuracy of CIFAR10 from 94.36\% to 93.76\%.
As a special case, the term inside summation can be turned into GELU when setting  $\mu = 1/\sqrt2$.
Further, the optimal pair of hyperparameters can be different for different tasks.

These two sections of Appendix lead to two conclusions: (i) $\alpha$ values are positively correlated with $\text{Var}(O(x) - m^*)$ and negative values of $\alpha$ help reduce the 3 variance terms on the right side of the equation; 
(ii) $\alpha$ values are negatively related to $\text{Cov}(x, o_c(x), o_a(x))$.
BETA regularization is not effective in this case as it keeps all $\beta$ values close. 
To increase the covariance term while satisfying condition (i), a zero-one loss-like approach is needed. 
SA-DARTS serves this purpose, keeping small $\alpha$ values to minimize variance and allowing for both large positive and negative values to differentiate relative contribution.


\subsection{Classification with Nas-Bench-201}

\begin{table*}[tpbh]
    \centering
    \resizebox{\textwidth}{!}{
    \begin{tabular}{|l|c|c|c|c|c|c|c|}
    \hline
        NAS Bench 201 & Cost & \multicolumn{2}{c}{Cifar 10} & \multicolumn{2}{c}{Cifar 100}  & \multicolumn{2}{c}{ImageNet}  \\
        & (hours) & Valid & 	Test	 &Valid &	Test &	Valid &	Test \\
        \hline
ResNet	&	N/A     & 90.83 &	93.91	& 70.50 &	70.89 &	44.10 &	44.23 \\
       optimal	&	N/A     & 91.61 &	94.37	& 73.49 &	73.51 &	46.77 &	47.31 \\
       \hline
DARTS (1st) \cite{liu2018darts} &	4       &  39.77$\pm$0.00  &	 54.3$\pm$0.00  &	15.03$\pm$0.00 &	15.61$\pm$0.00 &	16.43$\pm$0.00 &	16.32$\pm$0.00 \\
DARTS (2nd) \cite{liu2018darts} &	12	 & 39.77$\pm$0.00 &	54.3$\pm$0.00 &	15.03$\pm$0.00 &	15.61$\pm$0.00 &	16.43$\pm$0.00 &	16.32$\pm$0.00 \\
PC-DARTS \cite{xu2021partially}	      & 2     &	89.96$\pm$0.15 &	93.41$\pm$0.30 &	67.12$\pm$0.39 &	67.48$\pm$0.89 &	40.83$\pm$0.08 &	41.31$\pm$0.22 \\
SNAS \cite{xie2018snas} & 7 & 90.10$\pm$1.04 & 92.77$\pm$0.83 & 69.69$\pm$2.39 & 69.34$\pm$1.98 & 42.84$\pm$1.79 & 43.16$\pm$2.64 \\
DrNAS \cite{chen2020drnas}       &	4   &	91.55$\pm$0.00 &	94.36$\pm$0.00 &	73.49$\pm$0.00 &	73.51$\pm$0.00 &	46.37$\pm$0.00 &	46.34$\pm$0.00 \\
Beta-DARTS \cite{ye2022b}       &	4   &	91.55$\pm$0.00 &	94.36$\pm$0.00 &	73.49$\pm$0.00 &	73.51$\pm$0.00 &	46.37$\pm$0.00 &	46.34$\pm$0.00 \\
\hline
PC + Beta     &	2 &	91.53$\pm$0.00 &	94.22$\pm$0.00 &	73.13$\pm$0.00 &	73.17$\pm$0.00 &	46.32$\pm$0.00 &	46.48$\pm$0.00\\
SNAS + SA   & 7 &	91.55$\pm$0.00 &	94.36$\pm$0.00 &	73.49$\pm$0.00 &	73.51$\pm$0.00 &	46.37$\pm$0.00 &	46.34$\pm$0.00 \\
SA-DARTS       &	0.8  &	91.55$\pm$0.00 &	94.36$\pm$0.00 & 73.49$\pm$0.00 &	73.51$\pm$0.00 &	46.37$\pm$0.00 &	46.34$\pm$0.00 \\
SAC-DARTS        &	0.4  &	91.55$\pm$0.00 &	94.36$\pm$0.00 &	73.49$\pm$0.00 &	73.51$\pm$0.00 &	46.37$\pm$0.00 &	46.34$\pm$0.00 \\
    \hline
    \end{tabular}
    }
    \caption{Performance comparison on NAS-Bench-201 of our methods against existing SOTA models.}
    \label{tab:pcb-nb201-result}
\end{table*}

In addition to DARTS original search space, we also leverage NAS-Bench-201 to test the efficiency of our method.
Table~\ref{tab:pcb-nb201-result} summarizes the statistical comparison results on NAS-Bench-201 with 10 different \textit{random seeds} of SOTA methods.
Both SA-DARTS and SAC-DARTS can yield near-optimal performance with a faster convergence rate.
In addition, once the $\mu$ and $\nu$ combination Equation (7) is in the optimal neighborhood, the performance is not sensitive to the change in $\mu$ and $\nu$.
There exist several hyperparameters to achieve this result on NAS-Bench-201, one of such examples is $\lambda = \text{epoch}/2$, 
$\nu = 0.25$, 
and $\mu = $ \num{1e6}, 
mimicking $\mu \rightarrow \infty$, making a smooth approximation of Leaky ReLU.
Another SOTA hyperparameter is when $\nu = 1$, and $\mu$ can be any number. 
Also note that, although we give a budget of 50 searching epochs for all methods, our SA-DARTS and SAC-DARTS can always converge to the best architectures within 10 epochs.
Therefore, the actual convergence cost of SA-DARTS and SAC-DARTS is 0.8 and 0.4 days.
An advantage of SA is it can be easily implemented with DARTS-based methods.
For instance, our method can easily bring SNAS's previously sub-optimal performance to a SOTA state.

We present the CIFAR100 test accuracy of the SA-DARTS searched model (on NAS-Bench201) by replacing operators on the 1st edge of decreasing $\beta$ order: 73.5\%, 72.3\%, 70.78\%, 71.8\%, and 70.02\%.
This demonstrates SA-DARTS's ability to reflect the importance of operations through $\alpha$.
As shown in Table~\ref{tab:psnr_dlsrspace2}, our SA-DARTS is effective in that it can \textit{converge in only 0.4 days} under a 2 GPU day search budget, while Beta-DATRS cannot generalize well to super-resolution.


\subsection{Super-Resolution}\label{sec:app_sr}

\subsubsection{DLSR Search Space}

\begin{figure*}[tbph]
    \centering    \includegraphics[width=\textwidth]{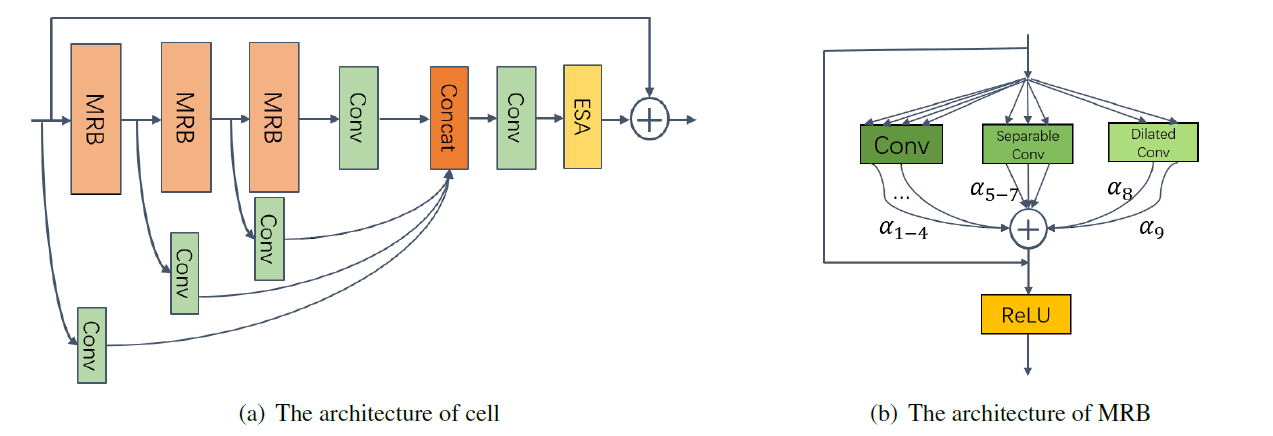}
    \caption{The macro and micro-level supernet during searching phase. The Cell-level Search Space of DLSR. The cell comprises 3 mixed residual blocks with an information distillation mechanism and an ESA block. The ‘Conv’ in (a) denotes the 1 x 1 convolution layer that cuts the channel number by half. (b) shows the architecture of mixed block, which is composed of multiple operations weighted by the parameter $\alpha$, residual skip-connection, and ReLU layer \cite{huang2021lightweight}.}
    \label{fig:dlsr}
\end{figure*}

The structure of DLSR \cite{huang2021lightweight} is shown in Figure~\ref{fig:dlsr}.
The macro space of DLSR is densely connected, and the micro space is carefully designed to enable mixed operations.
Although the supernet looks very different from the classification-based DARTS methods, given that DLSR adopts the DARTS for MRB blocks, we can still improve DLSR with various DARTS-variant toolbox.

\subsubsection{IMDN Search Space}

\begin{figure}[tpbh]
     \centering
     \begin{subfigure}[b]{0.49\textwidth}
         \centering
         \includegraphics[width=\textwidth]{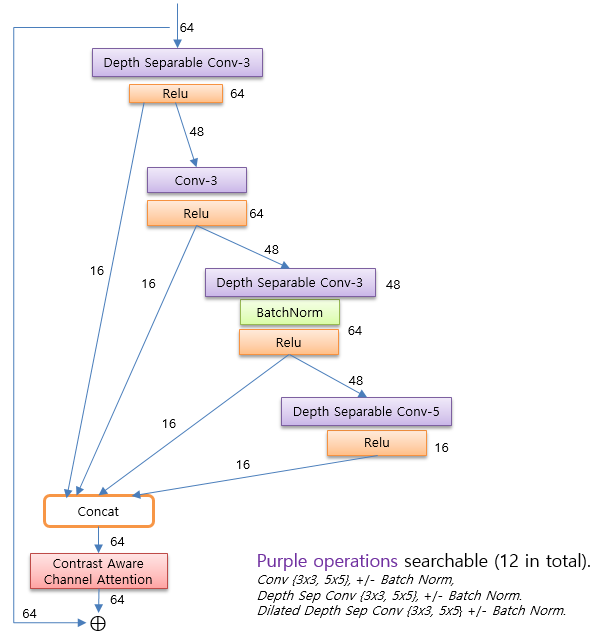}
         \caption{IMD block found by SA-DARTS.}
     \end{subfigure}
     \begin{subfigure}[b]{0.4\textwidth}
         \centering
         \includegraphics[width=\textwidth]{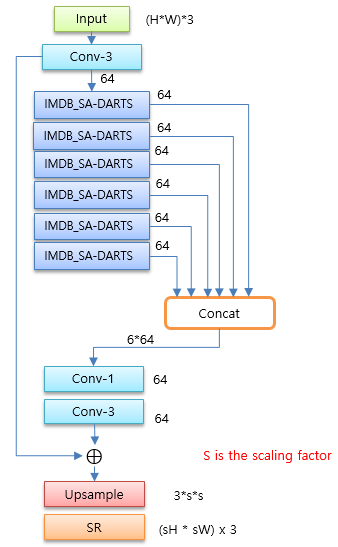}
         \caption{IMDN overall structure.}
     \end{subfigure}
        \caption{Improved IMDN model found by SA-DARTS}
        \label{fig:imdn-nas}
\end{figure}

As shown in Figure~\ref{fig:imdn-nas}, we search 4 opeartors in purple and try to replace the original conv3x3 by other operators.
The candidates are: conv, depth separable conv, dilated depth separable conv with kernal choice of 3x3 or 5x5, and choice of followed by batch norm. 
In total, 12 candidates are given.
Without FLOPs constraint, Beta-DARTS picked all conv5x5 operators, Figure~\ref{fig:imdn-nas}(a) shows final model searched by SA-DARTS. 
With FLOPs constraint, for Beta-DARTS, its genotype is: \textit{[depth sep conv3x3, conv3x3, depth sep conv3x3 + BN, depth sep conv5x5]}.
For SAC-DARTS, its genotype is: \textit{[dilated depth sep conv3x3, depth sep conv3x3, depth sep conv3x3, dilated depth sep conv3x3].}

\subsubsection{FLOPs Constraints}

\begin{table*}[t]
    \centering
    \resizebox{\textwidth}{!}{
    \begin{tabular}{lccccccc}
    \hline
       Method &		Search Cost (days) & Set5 &	Set14 &	B100 &	Urban100		& Params (K) &	Multi-Adds (G)\\
       \hline
DLSR &	 2+ &	37.91 &	33.56	&32.17 &	32.10 &	322.62 & 363.37	 \\
IMDN & - &	38.00 &	33.63&	32.19&	32.17&	715.18 & 654.53 \\
DLSR + Beta-DARTS &	 2.2 / 2.2  &	37.88&	33.46&	32.09&	31.71&	2243.00 &	2131.14\\
\hline
SA-DLSR (Ours) & 0.4 / 2.2  & 37.97 &	33.53 &	32.16 &	32.02 &	264.73	& 215.90 \\
SAC-DLSR (Ours) & 0.11 / 1.2 & 37.98 &	33.58 &	32.17 &	32.16 &	309.37 &	350.70 \\
\hline
IMDN + Beta-DARTS &	 2.2 / 2.2  &	\textbf{38.01} &	33.47&	32.11&	32.12&	1773.67 &	1745.04 \\
SAC-IMDN (Ours) & 0.7/1.2  & 37.98 &	\textbf{33.65} &	\textbf{32.20} &	\textbf{32.21} &	349.22	& 317.09 \\
\hline
IMDN + Beta-DARTS w/ FLOPs constraint &	 2.2 / 2.2  &	37.91 &	33.27&	32.05&	31.87 &	349.22  &	431.38  \\
SAC-IMDN w/ FLOPs constraint (Ours) &	 0.4 / 2.2  &	37.92 &	33.32&	32.08&	31.93 & 200.04	  &	293.89  \\
    \hline
    \end{tabular}
    }
    \caption{PSNR (x2 scale) results of different methods. Search is on DIV2K only, and evaluated on different SR Datasets with DLSR Search Space. The model searched by SA-DARTS achieves the best PSNR value among all methods with the DLSR search space, and can match IMDN \cite{hui2019lightweight} (different model space) with only half of the parameters.}
    \label{tab:psnr_dlsrspace2}
\end{table*}

Therefore, we applied FLOPs constraints on two search techniques to see if they can find smaller models with good performance.
We introduce a trade-off search for complexity as an additional loss term in addition to the L1 loss during search. 
The new loss is 
\begin{equation}
    L_{FLOPs} = \sum_i \frac{\beta_i c_i}{\sum c}
\end{equation}
where $c$ is the FLOPs calculated with 64 (for 1st block) or 48 (for rest blocks) channels on 1280x720 image,
$\beta$ is the architecture weight vector,
and $i$ is the index of candidate operators.
In the last two rows of Table~\ref{tab:psnr_dlsrspace}, we show that the models searched are indeed smaller, however, their performances slightly drop too.

\subsection{Comparing SA-DARTS with Other Regularizers}

As discussed in an example illustrated by FairDARTS in Section~\ref{sec:discrepancy} and shown in Figure~\ref{fig:darts_beta_mr_beta} (b), Beta-regularization suffers from competing issue of $\beta$ since its goal is to keep the change small and close to each other.
Under the weighted sum scheme, it is difficult to discern which operator is the best since top-2 operators contribute almost equally to the next feature map.
Moreover, we adjusted the softmax temperature for Beta-regularization, hoping that it would alleviate the discrepancy of discretization issue.
We have tried different original and Beta DARTS temperature values: {$0.1, 0.5, 1, 5, 10, 20$}.
Unfortunately, none of these values help differentiate the best candidates in Beta-regularization, as $\alpha$ of top-2 candidates are too close in all temperature settings.
We claim that the LSE loss used is a very rigid condition, and not enough exploration is given.
We empirically show that Beta-DARTS fails to generalize to other tasks with different data and model distributions.

The value of $\beta_{skip}$ remains the highest throughout the entire search process, and the ground truth candidate operation \textit{conv 3x3} fails to be selected.
It is partly due to the positive exponential calculation exploding the updating signal.
This intuition also partially explains why Beta-DARTS works: it uses LSE as a loss term and pushes all $\alpha$ to negative values.
Therefore, softmax of in Equation~\ref{eqn:beta} undergoes a minor change.
However, setting $\alpha$ values to negative does not entirely solve the problem.

\subsection{Negative Initialization}\label{sec:app_neg-darts}

SA-DARTS reduces all the $\alpha$ to negative numbers as shown in Figure~\ref{fig:mean_median_std} (b), and the leading privilege of $\beta_{skip}$ is constrained.
Our method is equivalent to shifting exponential calculation in Equation~\ref{eqn:beta} to a more stable output region, so the accumulation of skip-connection will not be exponentially exaggerated as the search continues.
The calculation of each operator's contribution $\beta$ in Equation (1) can cause unstable gradient updates, as positive $\alpha$ values can result in undesired larger changes when raised to the power of $e$, compared to negative $\alpha$ values.
In contrast, Equation (7) provides a smooth approximation of a linear activation function that allows for negative values and only increases linearly with respect to the absolute value. 
The advantage of SA-DARTS is that it not only constrains the $\beta$ update, but also brings stable standard deviation values of $\alpha$, as shown in Figure~\ref{fig:darts_beta_mr_beta} (c).
Thus, the probability updates are steadier, granting a stable search process.
As a result, SA loss is differentiable and negative $\alpha$ values allow for incremental gradient updates instead of exponential steps.
Next, we show that our SA regularization is necessary, rather than relying on negative $\alpha$ initialization.

To further validate that it is the SA regularization rather than the negative $\alpha$ that solves the skip dominance issue, we show that DARTS still suffers from skip aggregation with negative initialized $\alpha$.
Specifically, we initialize all $\alpha$ with $\num{1E-3} * \{-0.5,1,2,5\}$.
\begin{table}[tpbh]
    \centering
    \begin{tabular}{|c|c|}
    \hline
        Initialization &   Performance  \\
    \hline
        $-0.5$  & 54.30\%  \\
        $-1.0$ &  54.30\% \\
        $-2.0$    &  70.92\% \\
        $-5.0$ & 70.92\% \\
    \hline
    \end{tabular}
    \caption{Comparison of original DARTS with different negative initialized $\alpha$ values. Performance is reported at $50^{th}$ searching epoch on NAS-Bench-201 CIFAR10 test.}
    \label{tab:neg_init}
\end{table}

\subsection{Partial Channel Connection}

DARTS is memory inefficient because the supernet performs 8 operations for each edge.
Partial Channel DARTS (PC-DARTS) \cite{xu2019pc} has been one of few successful methods that can significantly speed up the DARTS with improved memory and computational efficiency.
PC-DARTS reduces the required run-time GPU memory by sampling a random subset of channels instead of sending all.
The channel sampling is done by a masking, which assigns 1 to selected channels and 0 to masked ones during both forward and backward step.

Rewriting the equation for mixed operation in DARTS, new feature map output $f_{i,j}^{PC}(x_i;S_{i,j})$ with channel sampling can be expressed as:
\begin{equation}
	 \sum_{o \in O}\frac{\exp(\alpha_o^{i,j})}{\sum_{o'\in O}\exp(\alpha_{o'}^{i,j})} o(S_{i,j}*x_i) + (1-S_{i,j})*x_i
\end{equation}

The masking portion $K$ is usually considered as a hyper-parameter, which is 4 for CIFAR10 and 2 for ImageNet.
The proportion of selected channel to be $1/K$ means a speed up of K times.
However, updating only portions of channels could introduce instability and fluctuation of the optimal connection.
Because, these architecture parameters are optimized by randomly sampled channels across iterations, and thus the optimal connectivity determined by them could be unstable as the sampled channels change over time.
To mitigate this issue, PC-DARTS assign learnable weights $q$ for each edge.
During the final selection, values of $p$ (softmax of $\alpha$) and $q$ are multiplied together as inputs for Equation~\ref{eqn:dartsargmax}.

\begin{equation}
	x_j^{PC} = \sum_{i < j} \frac{\exp{(\rho_{i,j})}}{\sum_{i' < j}\exp{\rho_{(i', j)}}}   f_{i,j}^{PC}(x_i;S_{i,j})
\end{equation}


\subsection{Why does SA-DARTS regularization work for Partial Channel? }\label{sec:mr-pc}

In this section, we study the partial channel connections with different regularization methods in a theoretical perspective.
Following \cite{xu2021partially}, we consider the example of searching $\mathcal{L}$ candidates and linear unit contains N neurons.
Let $\mathcal{E_{PC}}$ be the error function of partially-connected supernetworks.
\begin{equation}
\mathcal{E_{PC}} = \dfrac{1}{2} ( t - \sum_{l=1}^{L} \sum_{n=1}^{N} \beta_l w_{l,n}s_n x_n ) ^2 
\end{equation}

Here, $t$ is the target label, $x_n$ is the $n$th dimension of the input tensor, $w_{l,n}$ and $\beta_l$ are weights and softmax of architectural parameters. 
Similar to \cite{xu2021partially}, $s_n$ is a mask with binominal distribution of taking value of one with probability of $p_n$.
Thus, the original DARTS with no channel masking can be considered as one of the ensemble models with weighted sum of selected (with probability of $p_n$) and otherwise unselected channels.
The error function of ensemble model becomes $\mathcal{E}_{ens}$.
\begin{equation}
\mathcal{E}_{PC} = \dfrac{1}{2} ( t - \sum_{l=1}^{L} \sum_{n=1}^{N} \beta_l w_{l,n}s_n x_n )^2
\end{equation}

The expectation of $\delta \mathcal{E}_{PC} / \delta w_{l,n}$ associated with the random variable $s_n$ is
\begin{align} \label{eqn:epc-w}
    \mathbf{E}[\dfrac{\delta \mathcal{E}_{PC}}{\delta w_{l,n}}] &= 
    \dfrac{\delta \mathcal{E}_{ens}}{\delta w_{l,n}} +
    \sum_{z=1}^{L}   w_{z,n} \beta_l \beta_z \text{Var}(s_n) x_n^2 \\  
    & = 
    \dfrac{\delta \mathcal{E}_{ens}}{\delta w_{l,n}} +
    \sum_{z=1}^{L}   w_{z,n} \beta_l \beta_z p_n (1-p_n) x_n^2  
\end{align}

It is clear that $\mathcal{E}_{PC}$ is equivalent to $\mathcal{E}_{ens}$ with an additional term that brings model away from the ensemble model, thus driving the error term towards higher variance.

Similarly, the expectation of $\delta \mathcal{E}_{PC} / \delta \alpha_l$ can be found as:
\begin{align} \label{eqn:epc-a}
    \mathbf{E}[\dfrac{\delta \mathcal{E}_{PC}}{\delta \alpha_l}] &= 
    \dfrac{\delta \mathcal{E}_{ens}}{\delta \alpha_{l}} +
    \sum_{z=1}^{L} \sum_{n=1}^{N} \beta_z w_{z,n} w_{l,n}  p_n (1-p_n) x_n^2
\end{align}

It first suggests that partial channel technique introduces an additional quadratic regularization term when optimizing weights and the architecture parameters.
The regularization is expected to both reduce the overfitting.
To achieve this, the original PC method uses shared edge normalization to co-determine the connectivity of each edge, \textit{i.e.,} by multiplying the normalized coefficients of architecture and edge parameters.
Edge normalization avoids the unexpected accumulation of gradients led by mixed operation and undesired fluctuation led by channel sampling by penalizing the quadratic $w$ and reducing the variance of $s_n$.
As a result, PC plays a bigger role in reducing the error term in Equation~\ref{eqn:epc-a}.

On the other hand, with Beta-regularization, logsumexp is equivalent to a smooth max regularization term that keeps the additional terms in Equation~\ref{eqn:epc-w} and \ref{eqn:epc-a} from changing too much by penalizing excessively large $\beta$ terms.
However, as a result, all $\beta$ values are close to each other, reducing the additional regularization term to a simply weight regularization term scaled by channel sampling variance.
As shown in experiment session, PC-Beta can only achieve sub-optimal performances for all tasks we throw at it.

Further, our SA-DARTS not only minimizes mean values of architecture parameters and bringing all $\alpha$ values to large negative values, just as Beta-regularization, but also identifies the most contributing candidates.
With the linearly increasing weighing scheme, when supernet's weights are not adequately updated, the issue of dominating skip-connections discovered in \cite{zela2019understanding} is suppressed.
After the model weights are trained for some epoch and no longer prefers skip-connections, as shown in Figure~\ref{fig:darts_beta_mr_beta}(c), SA-DARTS promotes most contributing candidates to receive larger regularization effect, thus greater gradient signal.
Vice versa, SA-DARTS suppresses low-performing candidates to receive smaller gradient signals.


\subsection{More Analysis on Beta-DARTS \cite{ye2022b} on All Tasks}

\textbf{NAS-Bench-201:}
NAS-Bench-201 contains $5^6=15,625$ (4 nodes, 6 edges, and 5 candidate operations) trained architectures as a benchmark to allow researchers to focus on the quality of search methods without having to worrying about the retraining stage.
During our experiments, PC-DARTS often diverges too, while PC-Beta shows steady 90+ accuracy.

Let us take a step back and try to understand why PC-Beta only achieves sub-optimal performance on NAS-Bench-201 search space, and how does SA-DARTS and SAC-DARTS help with it.
Note that PC-DARTS selects a portion of channels to be updated in each iteration, which scarifies robustness for an improved speed.
To mitigate this issue, the edge normalization is necessary.
The current PC-Beta is sub-optimal in NAS-Bench201 (performance-wise) because 201 has deterministic edge numbers/connections. 
Unlike the DARTS framework that keep only 2 incoming edges for each node, in NAS-Bench201 we always have the same 6 edges from start to end, so edge normalization in PC does not have an effect. 
As a result, the performance fluctuation caused by random channel selection in PC is not mitigated by edge normalization.
We also observe that, although our side project PC-Beta performs better than PC-DARTS, it is sub-optimal compared to Beta-DARTS, because edge normalization of PC-DARTS does not have an effect on NAS-Bench-201.

\textbf{Super-Resolution:}
For Beta-regularization, unfortunately, the improvement is minimal, and the models are actually 5x bigger, as Beta-regularization converges to architectures with many conv 7x7 operations.
Recall that we criticized Beta-regularization in Section~\ref{sec:discrepancy}, pointing out that although it achieves good results on classification tasks, it does not mean that the $\alpha$ should stay close.
In the search space of DLSR, we clearly see that Beta-regularization tries too hard to find flatter loss landscapes, and it might mean an overfitting on the classification tasks.
This may due to the Beta-regularization looking to find the best region of architecture neighborhood, that any choice along $\alpha$'s gradient deviation can give a very similar performance, which we so-call the plateau region of neighbors.

\section{Acknowledgment and Future Directions}

This work is carrying a US Patent number US20240070455A1
 \cite{el2024systems}.
It has been used internally for Samsung projects including segmentation, super-resolution, denoising, etc \cite{kwon2024method}.
In future directions, this work is under investigation on depth estimation \cite{guo2021unsupervised, chen2020spatiotemporal}, multi-modal systems \cite{hor2024cm, el2025attentive}, language models \cite{ma2020improving, ma2020asking}, decision-making models \cite{zhou2018spiking, ma2019retailnet, deng2024stochastic, zhou2018developing, deng2023distributionally}, distributed agents \cite{peng2017distributed, peng2019distributed, pu2021server, zhou2022communication, zhou2019adaptive, zhou2020distilled, zhou2018design, zhoumodular}.
Interested readers are welcome to read our related papers above.

\end{document}